\documentclass[10pt,journal,compsoc]{IEEEtran}

\usepackage{cite}
\usepackage{graphicx}

\usepackage{amsmath, amsfonts}
\usepackage[cmintegrals]{newtxmath}

\usepackage[ruled]{algorithm2e}

\makeatletter
\newcommand{\removelatexerror}{\let\@latex@error\@gobble}
\makeatother

\usepackage[caption=false,font=normalsize,labelfont=sf,textfont=sf]{subfig}

\usepackage{cases}
\usepackage{threeparttable}
\usepackage{diagbox}
\usepackage{verbatim}
\usepackage{color}
\usepackage{multirow}

\newcommand{\tabincell}[2]{\begin{tabular}{@{}#1@{}}#2\end{tabular}} 

\begin{document}
% paper title
\title{AccEPT: An Acceleration Scheme for Speeding Up Edge Pipeline-parallel Training}

\author{Yuhao~Chen,~\IEEEmembership{Student Member,~IEEE,}
        Yuxuan~Yan,~\IEEEmembership{Student Member,~IEEE,}
        Qianqian~Yang,~\IEEEmembership{Member,~IEEE,}
        Yuanchao~Shu,~\IEEEmembership{Senior~Member,~IEEE,}
        Shibo~He,~\IEEEmembership{Senior~Member,~IEEE,}
        Zhiguo~Shi,~\IEEEmembership{Senior~Member,~IEEE,}
        and Jiming~Chen,~\IEEEmembership{Fellow,~IEEE,}
\IEEEcompsocitemizethanks{\IEEEcompsocthanksitem This paper was presented in part at the IEEE International Conference on Ubiquitous Communication (Ucom), Xi'an, China, July 2023\cite{yan2023edge}.
\IEEEcompsocthanksitem Y. Chen, Y. Yan, Q. Yang, S. He, Z. Shi, J. Chen are with the State Key Laboratory of Industrial Control Technology, Zhejiang University, Hangzhou 310027, China. (E-mail: \{csechenyh, yxyan44, qianqianyang20, ycshu, s18he, shizg, cjm\}@zju.edu.cn)}}%
% \thanks{Y. Chen, Y. Yan, Q. Yang, S. He, Z. Shi, J. Chen are with the State Key Laboratory of Industrial Control Technology, Zhejiang University, Hangzhou 310027, China. (E-mail: \{csechenyh, yxyan44, qianqianyang20, ycshu, s18he, shizg, cjm\}@zju.edu.cn)}}%

\IEEEtitleabstractindextext{%
\begin{abstract}
It is usually infeasible to fit and train an entire large deep neural network (DNN) model using a single edge device due to the limited resources. To facilitate intelligent applications across edge devices, researchers have proposed partitioning a large model into several sub-models, and deploying each of them to a different edge device to collaboratively train a DNN model. However, the communication overhead caused by the large amount of data transmitted from one device to another during training, as well as the sub-optimal partition point due to the inaccurate latency prediction of computation at each edge device can significantly slow down training. In this paper, we propose AccEPT, an acceleration scheme for accelerating the edge collaborative pipeline-parallel training. In particular, we propose a light-weight adaptive latency predictor to accurately estimate the computation latency of each layer at different devices, which also adapts to unseen devices through continuous learning. Therefore, the proposed latency predictor leads to better model partitioning which balances the computation loads across participating devices. Moreover, we propose a bit-level computation-efficient data compression scheme to compress the data to be transmitted between devices during training. Our numerical results demonstrate that our proposed acceleration approach is able to significantly speed up edge pipeline parallel training up to 3 times faster in the considered experimental settings.

\end{abstract}

\begin{IEEEkeywords}
Data compression scheme, distributed parallel edge training, adaptive quantization, bitwise compression
\end{IEEEkeywords}}

\maketitle
\IEEEdisplaynontitleabstractindextext
\IEEEpeerreviewmaketitle

\IEEEraisesectionheading{\section{Introduction}}
\label{sec:introduction}

\IEEEPARstart{E}{dge} computing has made it possible to run various deep learning (DL) applications such as computer vision\cite{redmon2018yolov3, sandler2018mobilenetv2}, natural language processing\cite{liu2021pre} and human activity recognition\cite{gu2021survey} on edge devices. These edge devices, like mobile phones, cameras, and watches, run the DL model locally with the raw data samples directly collected from the on-device sensors, providing real-time response as well as data privacy when compared with uploading the raw data to the cloud to run the model. Moreover, the raw data sampled by the edge device may be highly personalized. For example, for the human activity recognition task, the data distribution may vary for different individuals like heights and weights, or be greatly affected by environmental noise like weather changes, resulting in performance degradation of a pre-trained model. Hence, there are emerging demands of \textit{on-device training} to improve the performance of on-device inference and provide personalization. For example, Federated Learning(FL)\cite{konevcny2016federated}, a framework proposed by Google, is currently widely used for on-device training. However, due to the limited computation and storage capabilities, mobile devices are usually not able to train an entire DL model efficiently or even fail to run \cite{liu2013gearing}.

To tackle this issue, one of the solutions is the model partitioning technique, which partitions a DL model into several sub-models, and deploys them across a number of edge devices that execute the whole model collaboratively\cite{harlap2018pipedream, poirot2019split, chen2023ftpipehd, huang2019gpipe, vepakomma2018split}. With this technique, the data privacy is preserved since the raw data is only processed on the local device. The real-time response is also guaranteed by leveraging the computing resources of multiple edge servers. Moreover, compared with other techniques where the DL model is compressed to fit into a single edge device\cite{gordon2018morphnet, zeng2017mobiledeeppill, zhou2018rocket}, the model partitioning method also enjoys the advantage of running the original DL model such that preserves the model performance.

% To tackle this challenge, researchers have introduced model compression and model partitioning techniques for both the training and inference phases of the deep learning models. The model compression refers to compressing a large model into a smaller one by techniques such as model pruning\cite{gordon2018morphnet, you2019gate}, model quantization\cite{wu2016quantized, zeng2017mobiledeeppill} and knowledge distillation\cite{hinton2015distilling, zhou2018rocket}. The key idea of these methods is to reduce the number of model parameters or the number of bits needed to represent numerical values of the model weights. The model partitioning is to partition a DL model into several sub-models, and deploy them across a number of edge devices that execute the whole model collaboratively\cite{harlap2018pipedream, poirot2019split, chen2021ftpipehd, huang2019gpipe, vepakomma2018split}. Both the model compression and model partitioning approaches enable the edge devices to run DL models or parts of them locally, which provides data privacy and real-time response. The data privacy is preserved since the raw data is only processed on the local device. The real-time response is guaranteed by reducing the computation complexity in the model compression way, and leveraging the computing resources of multiple edge servers in the model partitioning way. 
% The model partitioning method also enjoys the advantage of running the original DL model instead of a compressed version such that preserves the model accuracy.

Most of the existing works on model partitioning consider \textit{on-device inference} by partitioning a model into two parts, with one part on the edge device and the other on the edge server\cite{kang2017neurosurgeon, ko2018edge, eshratifar2019jointdnn, laskaridis2020spinn, yao2020deep}. The edge device executes the first part of the model, and transmits the intermediate results to the server, which computes the remaining part, and returns the final output back to the device. However, few studies investigate model partitioning based on-device training. Some recent works proposed a learning framework called \textit{split learning}, which also splits a model into two parts and deploys one to the edge device and the other to the server \cite{vepakomma2018split, liu2020hiertrain, poirot2019split}. In this approach, however, the training process on the edge device is stalled until the backward gradients are received, which remarkably slows down the training process. 

\begin{figure}[htbp]
\centering
    \includegraphics[width=3.4in]{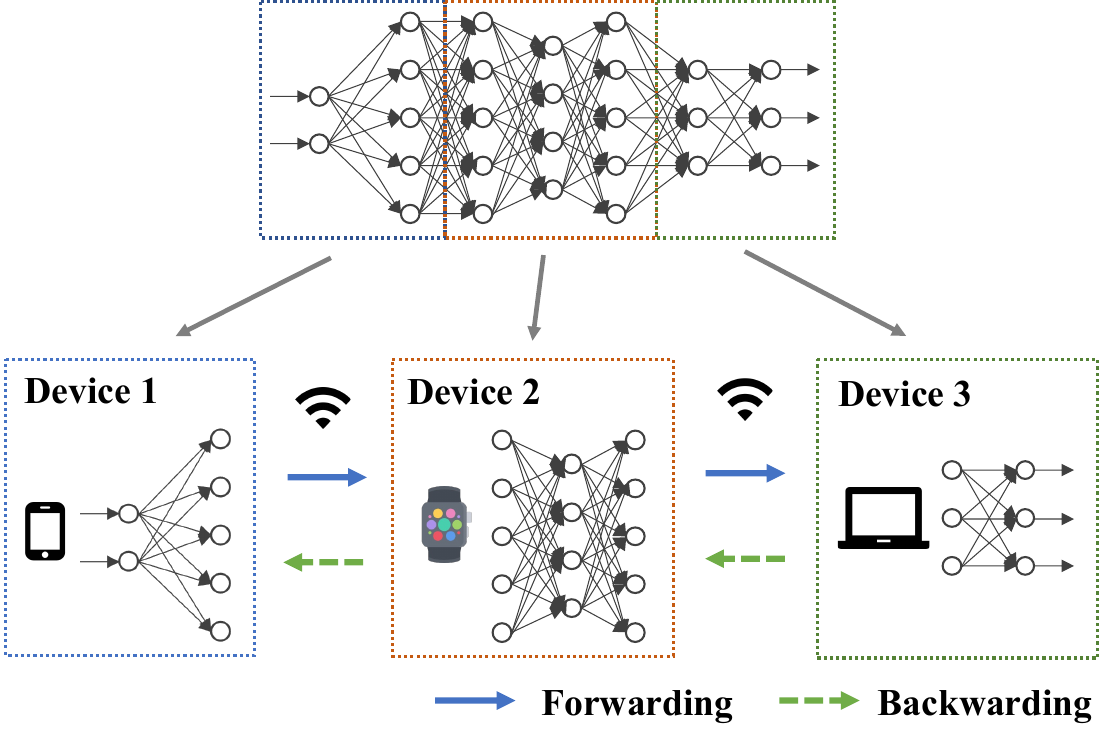}
\caption{An example of the distributed parallel edge training based on model partitioning.}
\label{model partitioning eg}
\end{figure}

To address this issue, in our previous work, we proposed an edge pipeline-parallel training approach called FTPipeHD\cite{chen2023ftpipehd}, which partitions a DL model into several sub-models and adopts the idea of the pipeline parallelism for GPU training acceleration to the on-device training. The pipeline mechanism avoids the stalling of devices by keeping feeding the data batches into the model without waiting for the newest weights to arrive, which achieves a significant speedup with a cost of slight performance degradation. Moreover, since edge devices have time-varying and heterogeneous computing resources, FTPipeHD also periodically estimates the computing capacity of each device during training. Based on this estimation, the DL model is dynamically partitioned as the training proceeds in order to accelerate training. Fig.~\ref{model partitioning eg} demonstrates the basic idea of the edge pipeline-parallel training, where a DL model is split into three sub-models and deployed across three edge devices.

Although FTPipeHD solves the stalling problem and provides a certain level of acceleration, there are still some factors that slow down training. Firstly, the latent features in the forward pass and the gradients in the backward pass, which need to be transmitted, can cause non-negligible communication latency during training. Unlike pipeline parallel training with multiple GPUs, where data is transmitted via high-speed cables, edge devices are generally interconnected by wireless links like WiFi and cellular networks, which usually have lower bandwidth compared with their wired counterparts.  Furthermore, for transmission through cellular networks, e.g., Long Term Evolution (LTE), 4G, and 5G, a large amount of data transmission may incur prohibitive costs charged by network operators, which may cause dissatisfaction among users of the edge devices. Secondly, the inaccurate estimation of the computing capacities of heterogeneous devices makes the model partitioning sub-optimal, thus slowing down training. As we will demonstrate in this paper, the execution time by each device estimated by FTPipeHD is highly inaccurate compared with the true execution time.

%In FTPipeHD, the minimum training time is achieved by calculating the optimal model partition point, where the layer-level execution time on each device should be estimated. The layer-level execution time refers to the execution time of a certain layer of the model. However, as we will demonstrate in this paper, the execution time estimated by FTPipeHD has a large error compared with the true execution time. The large estimation error will lead to a sub-optimal partition point, and hence the sub-optimal training time.

% Although the pipeline parallelism solves the stalling problem, the latent features in the forward pass and the gradients in the backward pass, which contain a large amount of data, need to be transmitted when training a batch. The transmission of these data may significantly slow down the training process. Unlike pipeline parallel training with multiple GPUs, where data is transmitted via high-speed links within one machine or through network cables across machines, edge devices are generally interconnected by wireless links like WiFi and cellular networks. These wireless links usually have lower bandwidth and network stability compared with their wired counterparts. Therefore a non-negligible network latency and even network failure should be considered in this training paradigm. Furthermore, for edge devices interconnected by cellular networks, e.g., Long Term Evolution (LTE), 4G, and 5G, a large amount of data transmission may incur prohibitive costs charged by network operators, which will cause dissatisfaction among users of the edge devices. 

To tackle the aforementioned issues, we introduce AccEPT in this paper, an acceleration scheme for \textbf{Acc}elerating \textbf{E}dge \textbf{P}ipeline-parallel \textbf{T}raining. Instead of predicting the layer-level execution time as in FTPipeHD, the scheme uses a light-weight latency predictor to accurately predict the sub-model's execution time by each device. To reduce the transmission latency, we propose a novel bit-level fast data compression scheme to effectively reduce the amount of transmission data. We implement AccEPT on edge devices and conduct comprehensive experiments. We summarize the main contributions of this paper as follows:

\begin{itemize}
\item We propose an acceleration scheme called AccEPT that speeds up the edge pipeline-parallel training, which consists of a light-weight latency predictor and a bit-level computation-efficient data compression scheme. The latency predictor provides an accurate sub-model's execution time prediction to better partition the model for training, while the compression scheme reduces the communication overhead.

\item We propose a light-weight latency predictor to dynamically estimate the sub-model's execution time by each device, which is pre-trained on data collected from various devices. During training, the proposed latency predictor performs continuous learning to adapt to unseen devices.

\item We propose a novel bit-level computation efficient data compression scheme that consists of an adaptive quantizer and a bit-wise encoder. The quantizer first quantizes the 32-bit floating-point data of the features and the gradients into lower-bit integer representation. Then the encoder is applied to the quantized data to reduce the redundant bits. A corresponding decompression scheme is also proposed.

\item We implement AccEPT on a set of commonly used edge devices and perform comprehensive evaluations to demonstrate the acceleration performance of AccEPT. Evaluation results show that AccEPT can speed up the edge pipeline training up to 3 times faster in the considered experimental settings.

\end{itemize}

The rest of this paper is organized as follows. Section \ref{sec: background and motivation} gives the background and motivation of our work. In section \ref{sec: system design}, we introduce the system overview and design details of AccEPT, followed by section \ref{sec: eval} where we present the numerical results to validate the performance of the proposed scheme. Finally, we conclude our work in section \ref{sec: conclusion}.

\section{Background and Motivation}
\label{sec: background and motivation}
\subsection{Pipeline-parallel Training}
\begin{figure}[htbp]
\centering
     \includegraphics[width=3.45in]{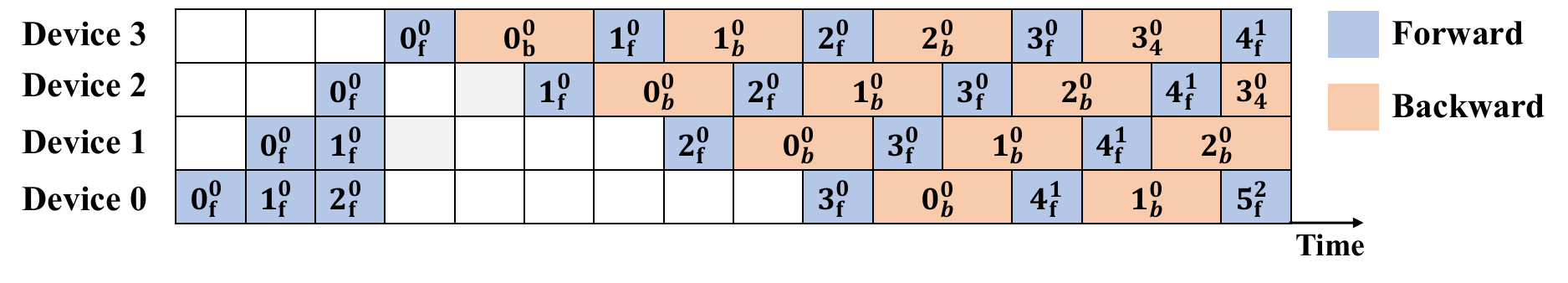}
\caption{The illustration of the pipeline parallel training across three devices.}
\label{pipeline example}
\end{figure}

We follow the idea of pipeline parallel training introduced in my previous work\cite{chen2023ftpipehd}, referred to as FTPipeHD, to enable distributed on-device training. By FTPipeHD, one device owns the raw data and manages the training, referred to as the \emph{central node}, while the other devices participating in the training are referred to as the \emph{worker nodes}. We present in Fig.~\ref{pipeline example} an example of this pipeline parallel training scheme with three devices, where device 1 is the central node, and $i^{ver}_\mathrm{f}$ and $i^{ver}_\mathrm{b}$ denote the forwarding and backwarding of the $i$-th training batch with the weights of the version number $ver$. We note that each device performs the forward and backward run of the local sub-model alternatively, which is referred to as \emph{one-forward-one-backward (1F1B)} rule. As illustrated in Fig.~\ref{pipeline example}, the consecutive forward and backward run by a device may be based on different versions of weights and different batches. Therefore, to guarantee the model convergence, the weights used to perform the forward run of a data batch are stashed for the later corresponding backward propagation of the same data batch. 

%Note that for a certain batch, the time of backwarding is longer than that of forwarding.

For instance, consider the training of the data batch 1, which is available at the central node. In traditional training without pipelining, the training of batch 1 starts only when device 1 receives the backward gradients of batch 0 and updates the weights accordingly, while by the pipelining scheme, the training of the batch 1 starts once the forward run of batch 0 is completed based on the initial weights of version 0 instead of the updated weights of version 1. We can observe from Fig.~\ref{pipeline example} that on device 1, although the weights have been updated to version 1 by the backward run of batch 0, batch 1 is still backwarded with the weights of version 0 to ensure that the same version of weights is used in the forward and backward pass of the same batch. We also note that for a certain batch, different versions of weights may be used on different devices, e.g., $5_\mathrm{f}^3$ on device 1, $5_\mathrm{f}^4$ on device 2, and $5_\mathrm{f}^5$ on device 3. This pipeline parallel training scheme aims to minimize the idle  time of devices during the training to accelerate the training.

% Moreover, since the computing and storage capacities of the edge devices are time-variant, the central device periodically collects the execution time from the worker devices, based on which the model is dynamically re-partitioned to minimize the total training time by solving a dynamic programming problem, which is elaborately formulated in our previous work~\cite{chen2021ftpipehd}. Note that the dynamic re-partitioning method makes the sub-model deployed on each device change periodically, hence the size of the transmission data varies overtime as well.

\subsection{The Large Amount of Transmission Data}
\label{subsec: large data size}
\begin{figure}[htbp]
\centering
     \includegraphics[width=3.45in]{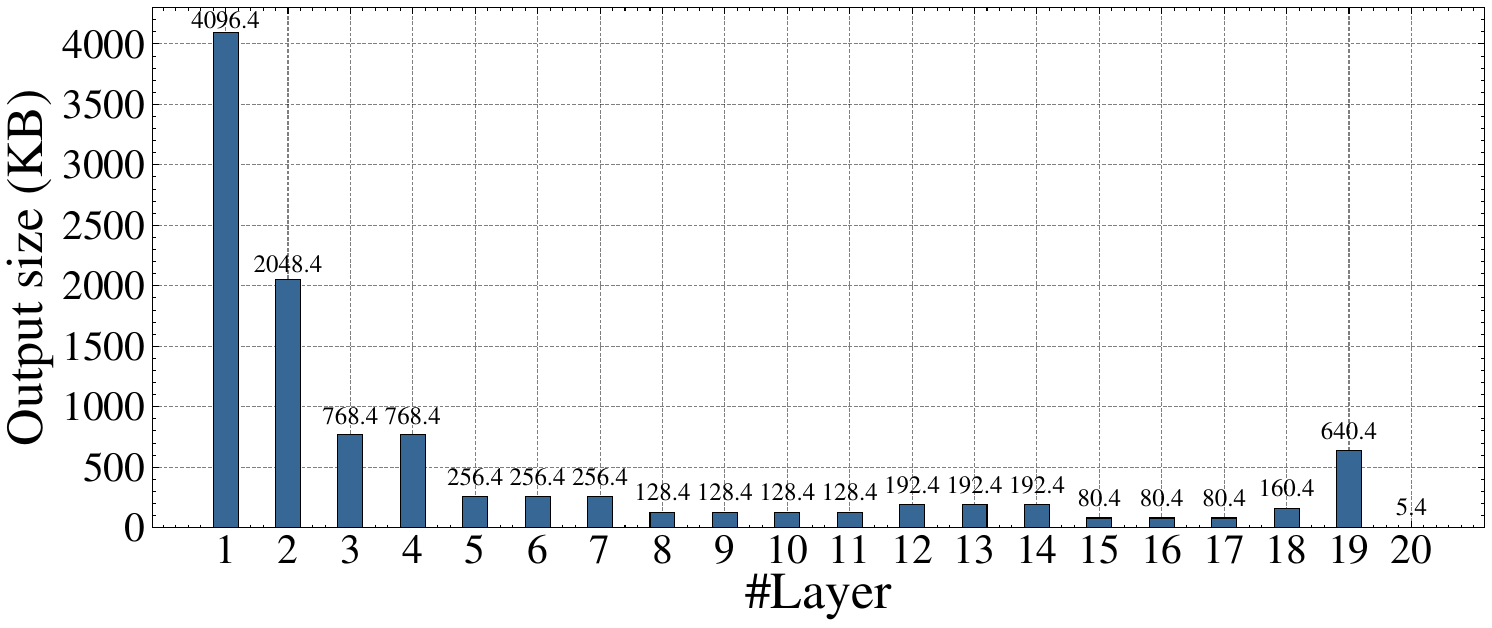}
\caption{Output size of each layer in MobileNetV2 when batch size is 128.} 
\label{model output size}
\end{figure}

Unlike model inference where the batch size of the input data is generally $1$, training a model usually requires a large batch size to provide computation efficiency and ensure convergence. In edge pipeline-parallel training, suppose that the model is partitioned at layer $l$ between devices $d_0$ and $d_1$, then the features and the gradients both of size $n \times w \times h \times c$, should be transmitted between two nodes, with a batch size of $n$, width of $w$, height of $h$ and a channel number of $c$. Each element is a 32-bit (4 bytes) floating-point value. Fig.~\ref{model output size} demonstrates the size of output of each layer $l$ in MobileNetV2 when batch size is 128 and the size of the input image is $3 \times 32 \times 32$. It can be observed that the largest output is about 4MB. If this is transmitted over the 4G cellular network whose bandwidth is 50Mbps on average, it still takes about 655ms to transmit it. Moreover, if the model is trained for 200 epochs and there are 391 batches in each epoch, about $4 * 200 * 391 \approx 312, 800 \mathrm{MB} \approx 305 \mathrm{GB}$ should be transmitted throughout the training, which may cause large data costs charged by network operators. To compress the transmitted data, researchers have proposed several data compression methods for the distributed inference based on model partitioning, where DNN-based compressor\cite{yao2020deep} can compress the data as much as possible while preserving the inference performance. However, for the training problem considered in this paper, compressors with large computation overhead can slow down the training of the whole model, and hence, are undesirable. Therefore, a computation-efficient compression scheme is needed to reduce the amount of data transmitted between nodes during the edge pipeline parallel training.

% suppose that we train the model for $M$ epochs, with $T$ batches in each epoch. During training, the data transmitted between two nodes, i.e., the features or the gradients, is $\boldsymbol{x} \in \mathbb{R}^{n \times w \times h \times c}$, with a batch size of $n$, width of $w$, height of $h$ and a channel number of $c$. Each element in $\boldsymbol{x}$ is a 32-bit (4 bytes) floating-point value. To complete the training, a total number of $M \times T \times n \times w \times h \times c \times 32 \times 2$ bits are transmitted between two nodes. Fig~\ref{model output size} demonstrates the output data size of each layer in MobileNetV2 when batch size is 128. 

\subsection{The Inaccurate Estimation of the Execution Time}
\begin{figure}[htbp]
\centering
     \includegraphics[width=3.45in]{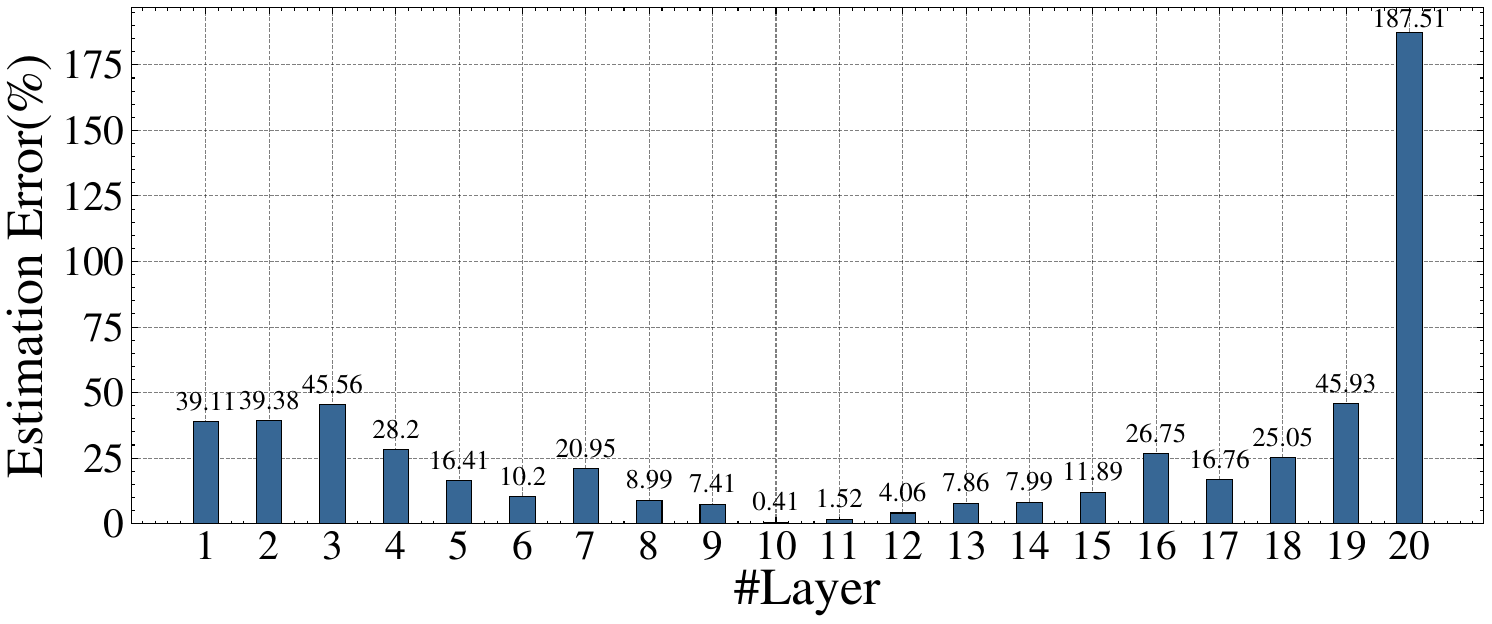}
\caption{Estimation error of the layer-level latency on $d_1$ for the 20-layer MobileNetV2.} 
\label{ftpipe time err}
\end{figure}

To minimize the training time, the layer-level execution time on each edge device should be accurately estimated so that the model can be optimally partitioned. In FTPipeHD, before the edge pipeline-parallel training, the central node $d_0$ profiles the total execution time of a forward pass and a backward pass for each layer, denoted as $T_{\mathrm{e}, j}^0$ for the $j$-th layer. The execution time of any sub-model on $d_0$ can be calculated by summing up the execution time of corresponding layers. During training, $d_0$ periodically collects the execution time of sub-models from each worker node. A ratio between the sub-model's execution time on the worker node and on the central node is calculated to represent the computing capacity of the worker node. Then for the $j$-th layer, its execution time $T_{\mathrm{e},j}^i$ on the worker node $d_i$ is estimated by multiplying this ratio with $T_{\mathrm{e},j}^0$. However, such an estimation is not accurate. Fig.~\ref{ftpipe time err} demonstrates the estimation error between the estimated and true layer-level latency on $d_1$ for a 20-layer MobileNetV2\cite{sandler2018mobilenetv2}, where the MobileNetV2 is partitioned by assuming all devices have the same computing capacity and $d_1$ is responsible for training the layers from 9 to 15. It can be seen from Fig.~\ref{ftpipe time err} that most of the estimation errors exceed $20\%$, where the estimation error of layer 20 is $187.51\%$. We also note that even though the overall execution time from layer 9 to 15 on $d_1$ is known by $d_0$, the estimation errors of layer 9 to 15 are still notable. The inaccurate estimation of execution time at each device leads to suboptimal model partitioning. Hence, it calls for a better prediction scheme in order to speed up the whole training process.

% This is due to the fact that layer 20 is a fully connected layer, leading to a shorter running time than the other convolutional layers, which induces a larger estimation error. 

\begin{figure*}[htbp]
\centering
    \includegraphics[width=7in]{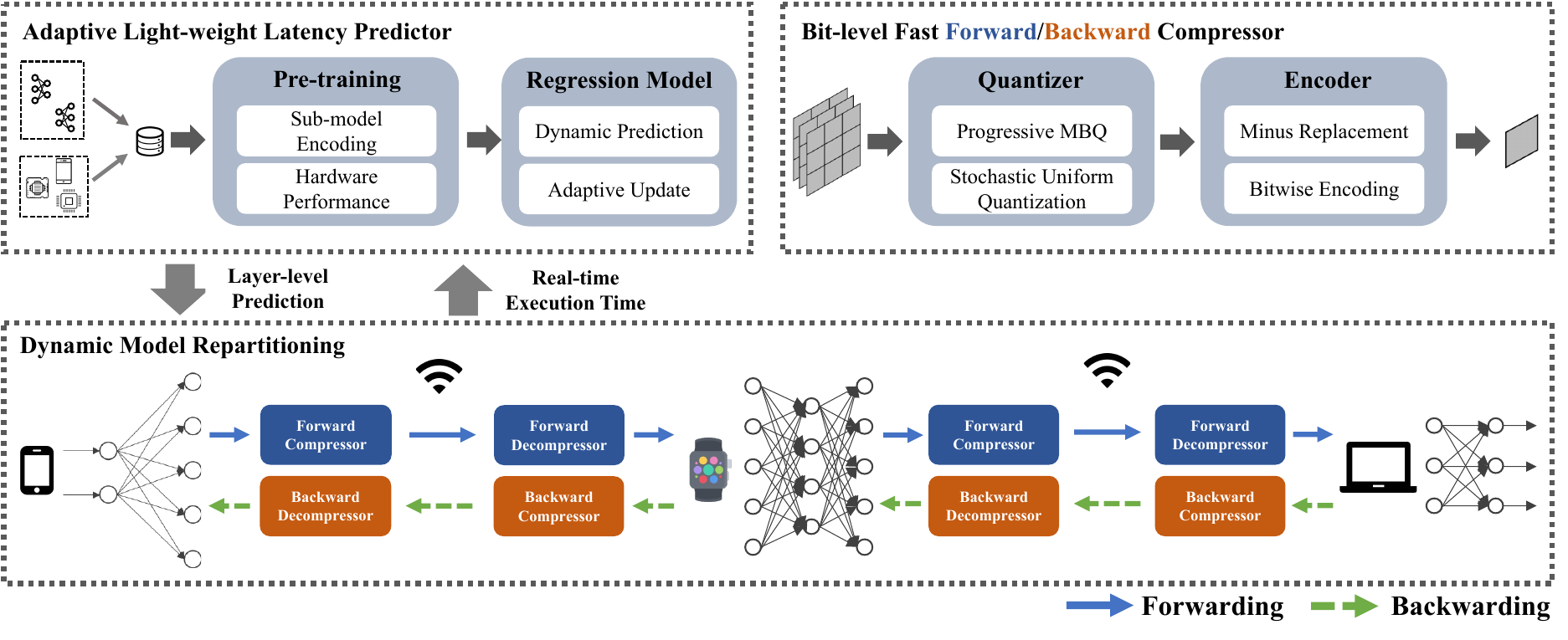}
\caption{The system overview of our proposed scheme.}
\label{system overview}
\end{figure*}

\section{System Design}
\label{sec: system design}
% In this section, we first introduce the system overview of the proposed compression and decompression scheme. Then we elaborate on the compressor and decompressor for both the forward and backward phases. For each compressor, we derive the transmission data size before and after compression, and calculate the corresponding compression ratio.
\subsection{System Overview}
Motivated by observations in section \ref{sec: background and motivation}, the proposed AccEPT scheme accelerates the edge pipeline-parallel training by accurately estimating the execution time for a better model partition point and reducing the amount of transmission data. To achieve this, AccEPT first pre-trains an adaptive light-weight latency predictor, which is a regression model containing only four fully-connected layers, with the data collected from various devices. During the training, the latency predictor estimates the execution time of each worker node and collects the exact execution time from unseen devices to update itself, which obtains a more accurate execution time estimation and thus a better partition point. Then, AccEPT uses a bit-level fast compression scheme to compress the data at each node before transmitting it to the next node. Specifically, it first quantizes the data to low-bit integers. Then it employs a bit-wise encoding to combine multiple low-bit integers into an 8-bit integer. At the receiver side, the transmitted data is decompressed in an inverse operation. On acquiring the accurate execution time and calculating the transmission time based on the reduced amount of transmitted data, AccEPT solves a dynamic programming problem to obtain the optimal model partition point, which aims to minimize the overall training time. Fig.~\ref{system overview} illustrates the overview of the proposed scheme.

% The system overview of our proposed scheme is illustrated in Fig.~\ref{system overview}. Based on the framework presented in Fig.~\ref{model partitioning eg}, we first partition a DNN model into several sub-models, each of which is deployed on a different edge device. We then train the sub-models in a pipeline parallel manner, which will be explained in detail in the sequel. The model is dynamically partitioned during the training process according to the real-time computing capacities and bandwidths of the participating devices. We apply a forward/backward compressor before the features/gradients are transferred to the next device in the forward/backward pass. During training, the original transmission data in floating-point representation, (i.e., the features output by a sub-model in the forward pass or the gradients in the backward pass) are first quantized to low-bit integers, with the features quantized by the proposed corrected MBQ method and gradients quantized by the stochastic uniform quantization, respectively. Then the proposed encoder further encodes the quantized data by the minus replacement and bitwise encoding approaches. When the compressed data is received by the next device, it is first decoded by a decoder that performs the reverse operation to the encoder. The decoded data is then dequantized into an approximation of the original transmission data, which is used for the training of the local sub-model.

\subsection{Adaptive Light-weight Latency Predictor}
\label{subsec: predictor}
Instead of estimating the layer-level execution time on each device, the proposed adaptive light-weight latency predictor provides an accurate sub-model's execution time of each worker and fast adaptation to unseen devices. It consists of three steps, i.e., the pre-training, dynamic prediction and adaptive update. Note that the latency predictor is specific to a certain model. All sub-models described below come from the same model.

\textbf{Pre-training.} Before the edge pipeline parallel training, we pre-train the latency predictor on the representative data collected from various devices. The latency predictor is a regression model with four fully-connected layers, which is defined as $T_{\mathrm{e},s}^i = f(s, h_i) $. The input of the latency predictor, namely $s$ and $h_i$, is the vector encoding of a sub-model and the vector characterizing the hardware performance of the worker node $d_i$, respectively. The output $T_{\mathrm{e},s}^i$ is the estimated execution time of the sub-model represented by $s$ on the worker node $d_i$.

In particular, we use the binary encoding to encode a sub-model. Suppose that the model being partitioned has $l$ layers, then $s$ is a vector of length $l$, with the $i$-th element representing the $i$-th layer. For each element in the encoding vector $s$, 1 means the presence of the corresponding layer in the sub-model, while 0 means the absence of the corresponding layer. For example, for a model with 10 layers, a sub-model which consists of its 3rd layer to 7th layer, the encoding of the sub-model $s$ is $[0,0,1,1,1,1,1,0,0,0]$. Note that the layer of a model can be a block consisting of several operators like the residual block in the MobileNetV2.

% Note that we suppose that the model can be linearly partitioned. Hence for the model which has a residual module like MobileNetV2, we treat the whole residual module as an inseparable layer. 

To characterize the hardware performance of each worker node, i.e., to obtain $h_i$, we use the execution time of ten representative sub-models on the worker node. Firstly, for a model with $l$ layers, there are totally $M=l(l+1)/2$ sub-models. We sort all these sub-models according to their FLOPs and choose ten sub-models whose indices satisfy $k*(M-1)/9, k=0,1,...,9$, as the representative sub-models. Then for an edge device, we put the execution time of the ten sub-models into a vector of length 10, which is $h_i$. Our experiments show that the process described above requires only a negligible amount of time.

To pre-train the regression model $f(s,h_i)$, we construct the training dataset $D = \{(s,h_i,T_{\mathrm{e},s}^i)\}$ by collecting data from various edge devices including edge servers, laptops and desktop personal computers (PC), whose specifications will be elaborated in the experiments. Note that the elements in the encoding vector are discrete, while those in the hardware performance vector are continuous. Hence, we use a dual-stream regression model inspired by Maple\cite{abbasi2022maple}. Specifically, the first stream processes the sub-model encoding vector through two fully-connected layers, the output of which will be concatenated with the hardware performance vector in the second stream and processed through additional layers. Finally, the output of the second stream is the predicted execution time of the sub-model.

\textbf{Dynamic Prediction.} During the edge pipeline parallel training, the latency predictor dynamically adjusts its predicted execution time based on the feedback of the exact execution time of each worker node, which will be introduced in the following. At the beginning of the edge pipeline parallel training, the central node $d_0$ requires all worker nodes to send $s$ and $h$ to $d_0$. After receiving $s_j$ and $h_j$ from $d_j$, the latency predictor calculates $T_{\mathrm{e},s}^j$. Similar to FTPipeHD\cite{chen2023ftpipehd}, each $d_j$ periodically sends its exact execution time of the local sub-model $s$ back to $d_0$ during the training. Different from FTPipeHD where this execution time is used to estimate the computing capacity of $d_i$, the latency predictor computes the difference between the received exact execution time and the estimated one. If the difference is larger than a pre-defined threshold $\epsilon$, it means that the computing resources of $d_j$ have changed substantially and $h_j$ should be recalculated. Based on the updated $h_j$, the execution time of the sub-model on $d_j$ is predicted again and thus the optimal model partition point is recomputed.

% Based on the newly constructed $h_i$, the optimal model partition point is recomputed. Since the sub-model on each node changes as the optimal model partition point changes, the weights on each layer should be redistributed, whose details can be referred to FTPipeHD.

\textbf{Adaptive Update.} To adapt to unseen devices encountered during the training, the proposed latency predictor will adaptively update itself. Specifically, the latency predictor collects the exact execution time $\hat{T}_{\mathrm{e},s}^i$ periodically sent by each worker node to construct a new dataset $\hat{D} = \{(s,h_i,\hat{T}_{\mathrm{e},s}^i)\}$, where $s$ and $h_i$ is sent to $d_0$ at the beginning of the edge pipeline parallel training. Then it performs the continuous learning to update itself using $\hat{D}$. As our experiment results show, even with eight samples, the proposed predictor can adapt well to unseen devices.

% To predict the execution time of a sub-model on The input of the latency predictor, namely $e$ and $s$, is the encoding of a sub-model and the vector characterizing the hardware performance \textcolor{red}{The definition of dataset. How to collect the dataset. How to define the representative set. The training time of the model}

% \textcolor{red}{Predict the latency based on two condition: seen / unseen devices. The condition where devices are unseen.}

% \textcolor{red}{How to adaptively update the model. The overhead and the training time of the update.}

\subsection{Bit-level Fast Compression Scheme}
\label{subsec: compressor}
To reduce the transmission latency between nodes, the proposed bit-level fast compression scheme compresses the data before transmission. The proposed scheme consists of an adaptive quantizer and a bit-wise encoder.

The adaptive quantizer aims to reduce the transmitted data size by transforming the floating-point data into a low $k$-bit integer representation. However, in the modern computer system, the minimum number of bits to store an integer is eight. Even the boolean type data is typically represented with eight bits in mainstream programming languages like C++ and Python due to the memory alignment requirements. Therefore, if $k$ is smaller than 8, each value of the transmitted data will contain $8-k$ redundant bits. One straightforward method is to concatenate several quantized values into one 8-bit integer. However, if 8 cannot be divided by $k$, there would still be some redundant bits. To solve this problem, we adopt multi-bit quantization (MBQ)\cite{lin2017towards} for the latent features, where the value is quantized into a linear combination of several one-bit values, making the quantized values easier to concatenate. For the gradients to be transmitted, we use the stochastic uniform quantization to balance the computation costs between the forward and backward pass as the calculation of gradients is usually more computation-heavy\cite{li2020pytorch}. Additionally, to reduce the transmitted data size as well as guarantee model convergence, the bit width $k$ is adaptively adjusted as the training proceeds.

After applying the adaptive quantizer to the transmitted data, the bit-wise encoder concatenates the quantized data along the batch size dimension due to the fact that batch size is typically
a power of 2. After encoding the quantized data, there would be no redundant bits. We introduce the forward quantizer, forward encoder, backward quantizer and backward encoder separately in the following.

\textbf{Forward Quantizer.} Similar to section \ref{subsec: large data size}, assume that the latent features $\boldsymbol{x} \in \mathbb{R}^{n \times w \times h \times c}$ is a four-dimensional tensor. We first obtain a vectorized form of $\boldsymbol{x}$, denoted by $\boldsymbol{x_\mathrm{v}} \in \mathbb{R}^{N \times 1}$, where $N=n \times w \times h \times c$. We then exploit MBQ\cite{lin2017towards} to quantize each element into a $k$-bit integer. More specifically, the quantized feature $\boldsymbol{x_\mathrm{v}}$ is approximated by the linear combination of $k$ binary bases, $\boldsymbol{\hat{x}_\mathrm{v}}$, given as,

% where an input value $r$ is approximated by a linear combination of coefficients $\boldsymbol{\alpha} = \{\alpha_1,...,\alpha_k\} \in \mathbb{R}$ and binary bases $\boldsymbol{b} = \{b_1,...,b_k\} \in \{-1, +1\}$, that is, $\hat{r} = \sum_{i=1}^{k}\alpha_i b_i$. Since each binary basis can be represented by only 1 bit, it provides great convenience for fully utilizing every bit of the transmitted data, which will be shown when introducing the bit-wise encoder.

\begin{equation}
 \boldsymbol{x_\mathrm{v}} \approx \boldsymbol{\hat{x}_\mathrm{v}} = \boldsymbol{B_\mathrm{v} \alpha},
\label{MBQ}	
\end{equation}

\noindent where $\boldsymbol{B_\mathrm{v}} \in \{-1, +1\}^{N \times k}$ is a binary matrix, and $\boldsymbol{\alpha} \in \mathbb{R}^{k \times 1}$ is the coefficient vector. Since every element of $\boldsymbol{B_\mathrm{v}}$ is only 1 bit, eight of them can be combined into an 8-bit integer, which allows us to do bit-wise encoding easily. The optimal $\boldsymbol{B^*}$ and $\boldsymbol{\alpha^*}$ are obtained by minimizing the quantization error:

\begin{equation}
\boldsymbol{B_\mathrm{v}^*}, \boldsymbol{\alpha^*} = \mathop{\arg\min}\limits_{\boldsymbol{B_\mathrm{v}},\boldsymbol{\alpha}}\Vert \boldsymbol{x_\mathrm{v}} - \boldsymbol{B_\mathrm{v} \alpha} \Vert ^2 .
\label{minimizing error}	
\end{equation}

However, we note that the optimization problem given in \eqref{minimizing error} is NP-hard\cite{qu2020adaptive}. Hence, we calculate the $\boldsymbol{B_\mathrm{v}^*}$ and $\boldsymbol{\alpha^*}$ alternatively at each epoch, following the idea in \cite{zhang2018lq}. When training batch 0 of every training epoch, we initialize the $\boldsymbol{B^0_\mathrm{v}}$ and $\boldsymbol{\alpha^0}$ by the residue-based method proposed in \cite{guo2017network}. In this residue-based method, we iteratively calculate them by minimizing the error between the residue and the linear combination $\boldsymbol{B^0_\mathrm{v}}\boldsymbol{\alpha^0}$, where the residue is defined as the error between the original data and the $\boldsymbol{B^0_\mathrm{v}}\boldsymbol{\alpha^0}$ of the previous iteration. Then, we denote the vectorized form of the intermediate feature as $\boldsymbol{x_\mathrm{v}^t}$, and the set of all the possible linear combinations of a given $\boldsymbol{\alpha}^t$ and binary bases as $\mathbb{P}^t$, where $t$ refers to the batch $t$. For example, if $k=2$, $\boldsymbol{\alpha}^t=(\alpha_1, \alpha_2)$, we have $\mathbb{P}^t = \{-\alpha_1-\alpha_2, \alpha_1 + \alpha_2, -\alpha_1+\alpha_2, \alpha_1+\alpha_2\}$. When training batch $t$, we first determine the $\boldsymbol{B_\mathrm{v}^t}$ by finding the value in $\mathbb{P}^{t-1}$ that is closest to each value in $\boldsymbol{x_\mathrm{v}^t}$, which can be done by sorting $\mathbb{P}^{t-1}$ and performing element-wise binary search on each value in $\boldsymbol{x_\mathrm{v}^t}$. This can be further speeded up by tensor-level operation that treats $\boldsymbol{x_\mathrm{v}^t}$ as a tensor in practical implementation enabled by coding libraries like Pytorch or Tensorflow. Experiments show that the tensor-level search is 10 times faster than the element-wise binary search. With $\boldsymbol{B^t_\mathrm{v}}$ determined, we then update $\boldsymbol{\alpha}^t$ as follows, which is based on the closed-form solution to \eqref{minimizing error}:

\begin{equation}
 \boldsymbol{\alpha}^{\mathrm{cur}} = (\boldsymbol{{B^t_\mathrm{v}}}^{\mathrm{T}}\boldsymbol{B_\mathrm{v}^t})^{-1}\boldsymbol{B^t_\mathrm{v}}^{\mathrm{T}}\boldsymbol{x_\mathrm{v}^t},
\label{alpha solution}	
\end{equation}

\begin{equation}
 \boldsymbol{\alpha}^t = \beta\boldsymbol{\alpha}^{t-1} + (1-\beta)\boldsymbol{\alpha}^{\mathrm{cur}},
 \label{alpha update}	
\end{equation}

\noindent where $\beta = \mathop{\min}(0.9, (1 + t)/(10 + t))$. 

As shown in \cite{nagel2019data}, quantization may introduce the biased error to the data, which distorts the distribution of the original data, and may hurt the convergence of the training. Hence, inspired by \cite{nagel2019data}, we further introduce a bias correction to the quantized data. Note that the quantization can be viewed as a noise added to the original data, i.e., $\boldsymbol{\hat{x}} = \boldsymbol{x} + n_\mathrm{q}$, we have 

\begin{equation}
\begin{aligned}
\mathbb{E}[\boldsymbol{\hat{x}}] &= \mathbb{E}[\boldsymbol{x} + n_\mathrm{q}] \\ 
&= \mathbb{E}[\boldsymbol{x}] + \mathbb{E}[n_\mathrm{q}],
 \label{biased error}	
 \end{aligned}
\end{equation}

\noindent where we observe that we can subtract the mean of the quantization error $n_\mathrm{q}$, i.e., $\mathbb{E}[n_\mathrm{q}]$, from the quantized data to eliminate the biased error. Note that in our proposed scheme, the dequantization is performed on the next device. Therefore, the forward compression scheme calculates the $\mathbb{E}[n_\mathrm{q}]$ and transmits it to the next device, which is then subtracted from the dequantized data.

The bit width $k$ is adaptively adjusted as the training proceeds. At the beginning of the training, we use a small $k$, which is 2, to reduce the communication latency, such that accelerating the convergence of the DNN model. We increase the $k$ by 1 every time the learning rate of the model is altered at a certain epoch according to the optimizer strategy, up to an upper limit of 4 bits. We note a larger bit width reduces the quantization error, and hence helps the convergence of the model in the later stage of training. Experimental results demonstrate that the model accuracy of the proposed adaptive bit width outperforms that of the fixed 2, 3 and 4 bit width and even the one without quantization.

Note that in DNN model training, all the operations in the forward pass should be differentiable so that the gradients can be calculated in the backward pass. The quantization operation, however, is not differentiable. To tackle this issue, we adopt the generally-used straight through estimator (STE)\cite{bengio2013estimating} to approximate the gradients of the quantization operation for simplicity. Specifically, suppose that $\boldsymbol{\hat{x}_\mathrm{v}} = q(\boldsymbol{x_\mathrm{v}})$ and $I(\cdot)$ is the identity function, then in the backward pass, we have

\begin{equation}
\frac{\partial \boldsymbol{\hat{x}_\mathrm{v}}}{\partial \boldsymbol{x_\mathrm{v}}} = \frac{\partial q(\boldsymbol{x_\mathrm{v}})}{\partial \boldsymbol{x_\mathrm{v}}} \approx \frac{\partial I(\boldsymbol{x_\mathrm{v}})}{\partial \boldsymbol{x_\mathrm{v}}} = 1.
\end{equation}

%The STE technique uses the identity function as the derivative of the rounding function.

\begin{figure}[htbp]
\centering
   \includegraphics[width=3.2in]{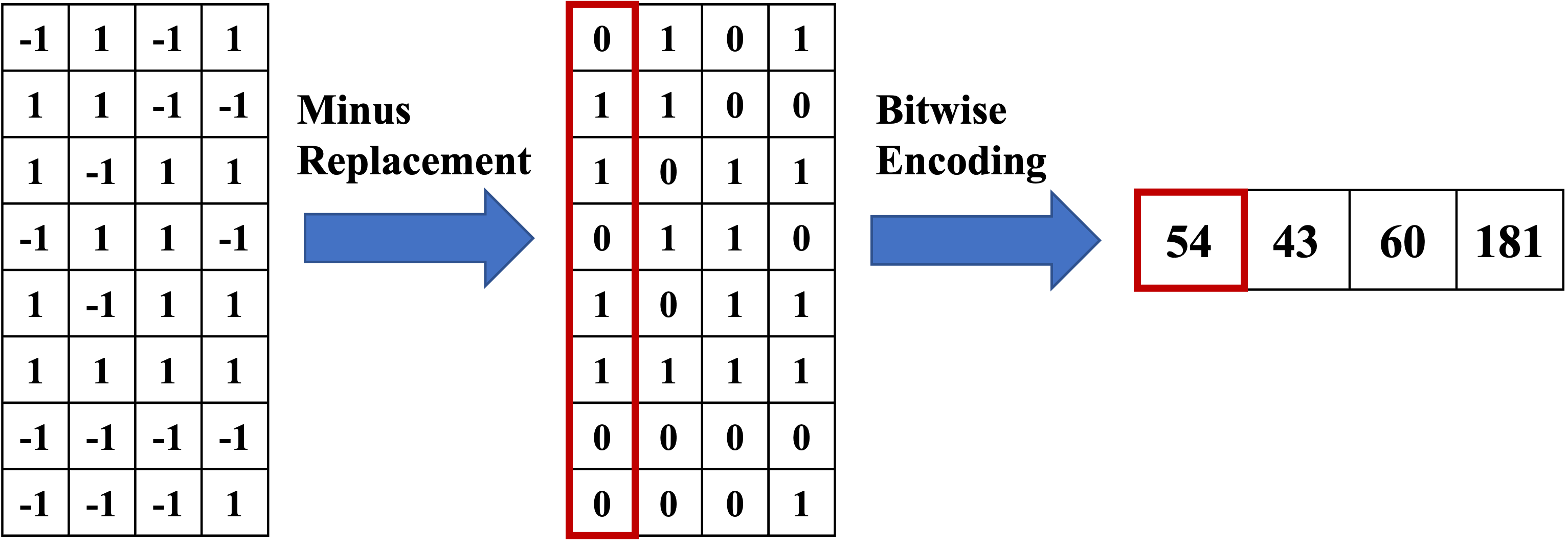}
\caption{An example of the forward encoder.}
\label{compressor}
\end{figure}

\textbf{Forward Encoder.} After quantizing the latent features, the forward encoder concatenates the binary bases along the first dimension. Firstly, since -1 requires at least 2 bits to represent, we replace all the -1s in binary bases with 0 before encoding. We then concatenate eight elements of $\boldsymbol{B}$ into one 8-bit integer. More specifically, the binary vector $\boldsymbol{B_\mathrm{v}}$ is first reshaped back to $\boldsymbol{B}$ of size $n \times w \times h \times c \times k$. We then combine every consecutive eight binary elements along the first dimension into an 8-bit integer, which results in a tensor of size $\frac{n}{8} \times w \times h \times c \times k$, denoted by $\boldsymbol{B_\mathrm{c}}$. Fig.~\ref{compressor} gives a simple example of encoding an $8 \times 1 \times 1 \times 1 \times 4$ binary tensor into an $1 \times 1 \times 1 \times 1 \times 4$ one. In this example, we first replace all the -1s with 0s. Then in the red-boxed part, eight binary numbers are encoded into one 8-bit integer, that is, $(00110110)_2 = 54$. Moreover, the proposed encoding method can be viewed as an encryption process, which enhances data privacy to a certain extent.

\textbf{Backward Quantizer.} For the gradients to be transmitted, we aim to quantize the floating-point gradients $\boldsymbol{g} \in \mathbb{R}^{n \times w \times h \times c}$ to $\boldsymbol{g_\mathrm{q}} \in \mathbb{Z}^{n \times w \times h \times c}$ in $k$-bit integer representation. More specifically, we first scale the $\boldsymbol{g}$ into the range $[-2^{k-1}, 2^{k-1} - 1]$ by dividing it with the scaling factor $s = \frac{c}{2^{k - 1} - 1}$ with $c = \mathbb{\max}\vert \boldsymbol{g}\vert$. Next, we employ stochastic rounding instead of the nearest rounding, which is a more common rounding technique. This is due to the fact that with the nearest rounding, small gradients are always rounded to zero, which may cause gradient vanishing. Instead, the stochastic approach rounds the data upwards or downwards with a certain probability, i.e.,
\begin{subequations}
\begin{numcases}{\boldsymbol{g_\mathrm{q}} =}
	\lfloor \boldsymbol{g} / s 	\rfloor , & w.p. $1 - pr$ \\
	\lfloor \boldsymbol{g} / s \rfloor + 1 , & w.p. $pr$,
\end{numcases}
\label{bitwise decompress}	
\end{subequations}

\noindent where $pr = \boldsymbol{g} / s- \lfloor\boldsymbol{g} / s \rfloor$. Numerical analysis\cite{gupta2015deep} has also revealed that stochastic rounding is an unbiased rounding scheme. 

The bit width is also adaptively adjusted in the backward quantizer as follows. Initially, we set bit width $k$ to 8, which then drops to 4 when the learning rate changes. Note that the quantization bit width of the gradients reduces as the training proceeds in contrast to the case in the forward compression. This is due to the fact that at the beginning of training, gradients change more significantly so that more accurate gradients are needed for the model to learn. As the training proceeds, the gradients get smaller when the model converges. Hence, a small bit width is sufficient.

\textbf{Backward Encoder.} Similar to the forward encoder, to make full use of the 8 bits of an integer, we encode the quantized gradients $\boldsymbol{g_\mathrm{q}}$ into $\boldsymbol{g_\mathrm{c}}$. If the bit width $k$ equals 8, the $\boldsymbol{g_\mathrm{q}}$ will be directly transmitted without encoding, i.e., $\boldsymbol{g_\mathrm{c}} = \boldsymbol{g_\mathrm{q}}$. For $k = 4$, we add 8 to each value of $\boldsymbol{g_\mathrm{q}}$, which moves the data range from $[-2^{4-1}, 2^{4-1} - 1]$ to $[0, 15]$. Then similarly, we combine two tensors along the first dimension into one tensor. In this case, the quantized gradients $\boldsymbol{g_\mathrm{q}}$ reduces to $\boldsymbol{g_\mathrm{c}}$ of size $\frac{n}{2} \times w \times h \times c$.

\textbf{Transmission Data Comparison.} Now we compare the size of transmission data before and after the compression scheme. For the forward run, before compressing the features, the latent features $\boldsymbol{x} \in \mathbb{R}^{n \times w \times h \times c}$ to be transmitted are in 32-bit floating-point representation, with a total size of $(n \times w \times h \times c \times 32)$ bits. After compression, the transmission data is comprised of the coefficient vector $\boldsymbol{\alpha} \in \mathbb{R}^{k \times 1}$, the encoded binary bases $\boldsymbol{B_\mathrm{c}} \in \mathbb{Z}^{\lceil \frac{n}{8} \rceil \times w \times h \times c \times k}$ and the mean of the quantization error $m_\mathrm{q} \in \mathbb{R}$, where the total size equals $(32k + \lceil \frac{n}{8} \rceil \times w \times h \times c \times k \times 8 + 32)$. The compress ratio $c_\mathrm{f}$ is given by 

\begin{equation}
\begin{aligned}
c_\mathrm{f} = \frac{n \times w \times h \times c \times 32}{\lceil \frac{n}{8} \rceil \times w \times h \times c \times k \times 8 + 32(k+1)}.
 \label{compress ratio}	
 \end{aligned}
\end{equation}

\noindent Note that if the size of $\boldsymbol{x}$ is large enough and $n$ is usually a multiple of 8, the compression ratio is approximately $\frac{32}{k}$. Similarly, for the gradients, the compressed transmission data contains the encoded gradients $\boldsymbol{g_\mathrm{c}}$ and a 32-bit scaling factor $s \in \mathbb{R}$, with the total size being $\frac{n}{8/k} \times w \times h \times c \times 8 + 32$, where $k$ equals 4 or 8. Thus the compress ratio $c_\mathrm{b}$ is also approximately $\frac{32}{k}$. 

\subsection{Bit-level Fast Decompression Scheme}
After encoding the quantized data, the data is transmitted to the next edge device, where the received data is then decompressed by the bit-level fast decompression scheme that performs the reverse operation to the compression. The decompression consists of a decoder and a dequantizer. Similarly, we introduce the forward decoder, forward dequantizer, backward decoder and backward dequantizer separately.

\textbf{Forward Decoder.} To decode the received $\boldsymbol{B_\mathrm{c}}$, we extract eight 1-bit values from each 8-bit integer of $\boldsymbol{B_\mathrm{c}}$. We then perform the minus recovery, which replaces all the 0s to -1 to obtain the original binary bases $\boldsymbol{B}$. 

\textbf{Forward Dequantizer}. After decoding, we have the coefficient vector $\boldsymbol{\alpha} \in \mathbb{R}^{k \times 1}$, the binary bases $\boldsymbol{B} \in \{-1, +1\}^{n \times w \times h \times c \times k}$ and the mean of the quantization error $m_\mathrm{q} \in \mathbb{R}$. The dequantized data $\boldsymbol{\hat{x}}$ can be simply obtained by matrix multiplication $\boldsymbol{B_\mathrm{v} \alpha}$, where $\boldsymbol{B_\mathrm{v}}$ is the vectorized form of $\boldsymbol{B}$. We further subtract $\boldsymbol{\hat{x}}$ by $m_\mathrm{q}$ to correct bias as mentioned in section~\ref{subsec: compressor}, which is then fed into the local sub-model to perform the forward run at the current worker node.

\textbf{Backward Decoder.} If the current bit width is 8, the decoding phase is skipped. For $k = 4$, the gradient decoder decodes the $\boldsymbol{g_\mathrm{c}}$ simply by dividing each 8-bit integer into two 4-bit integers and the minus values in $\boldsymbol{g_\mathrm{q}}$ are then recovered by subtracting each element by 8.

\textbf{Backward Dequantizer}. The $\boldsymbol{g_\mathrm{q}}$ is dequantized to $\boldsymbol{\hat{g}}$ by multiplying it with the scaling factor $s$, i.e., $\boldsymbol{\hat{g}} = \boldsymbol{g_\mathrm{q}} * s$. The decompressed gradients $\boldsymbol{\hat{g}}$ is an approximated version of the original gradients $\boldsymbol{g}$, both of which are in floating-point representation. The $\boldsymbol{\hat{g}}$ is then used for back propagation on the current device.

\subsection{Optimal Model Partitioning}
\label{subsec: optimal model partitioning}
After accurately predicting the execution time on each node and reducing the transmitted data size between nodes, we accelerate the edge pipeline-parallel training by finding the optimal model partition point. Following FTPipeHD, we calculate the optimal model partition point by solving a dynamic programming (DP) problem. Let $A(L,N)$ denote the optimal time used by the slowest node among $N$ nodes to collaboratively train a $L$-layer model. We start with $N=1$, i.e., only $d_0$ participate in the training, that is, 

\begin{equation}
A(l, 1) = T_\mathrm{e}^0(1, l),  \forall l \in [1:L],
\end{equation}

\noindent where $T_\mathrm{e}^0(a,b) = \sum\limits_{j=a}^{b}T_{\mathrm{e},j}^0$. Then we find $A(L,N)$ by iteratively calculating the following at each step:

\begin{align}\label{dp}
A(l, n) = \min\limits_{1\leq p < l}\max \begin{cases}
		A(p, n-1) \\ 
		2 \times T_{\mathrm{c},p}^{n-2} \\
		T_\mathrm{e}^{n-1}(p + 1, l)
	\end{cases} \forall l\in [n:L] ,
\end{align}
where $n$ is increased by one at each iteration until $n=N$, and $T_{\mathrm{c},j}^{i}$ denotes the communication time of sending the output of the $j$-th layer from $d_i$ to $d_{i+1}$. After calculating $A(L,N)$, the optimal partition point is then constructed by a backtracking method, which can be referred to FTPipeHD for more details. When calculating $A(l, n)$, we note that there are two main differences compared with the approach used in FTPipeHD. Firstly, the $T_\mathrm{e}^{n-1}(p + 1, l)$ in AccEPT is calculated by the latency predictor and more accurate, while it is calculated by the estimated computing capacity in FTPipeHD. Secondly, we reduce the transmission data with the proposed data compression scheme, leading to a smaller $T_{\mathrm{c},j}^{i}$ used in this work. As our experiments show later, the proposed acceleration scheme can acquire a better model partition point and further reduce the total training time.

\section{Evaluation}
\label{sec: eval}
\subsection{AccEPT Implementation}
We implement the proposed AccEPT on off-the-shelf devices with Pytorch of version 1.4.0. Note that any version higher than 1.4.0 may not be feasible due to the lack of support for reloading the weights during the training, which is necessary for the pipeline parallel training. We employ the commonly used lightweight Flask of version 1.1.2 to transmit data between devices. The implementation of the asynchronous pipeline-parallel training uses a condition lock to strictly follow the 1F1B rule.

% \subsection{Training Setup}
% We choose two representative DL applications to evaluate the performance of AccEPT, i.e., the image classification task and the human activity recognition task. For the image classification task, we use the CIFAR10 dataset and train a MobileNetV2 model\cite{sandler2018mobilenetv2}. For the human activity recognition task, we use the UCI-HAR dataset\cite{anguita2013public} collected from smartphone's inertial sensors and train a simple five-layer Convolutional Neural Network (CNN) model proposed in \cite{chen2020fedhealth}. 
% For the evaluation of the performance of our latency predictor, We gather data from diverse devices, including CPUs such as Intel(R) Xeon(R) CPU E5-2640, AMD EPYC 7402, and Huawei Kunpeng 920, as well as mobile devices like Samsung Galaxy A9. To limit resource usage and achieve more precise experimental results, we employed the taskset tool to control the number of processor cores executing tasks on some devices, thus simulating various levels of computing capacity. We measure the latency of all possible sub-models on every device, yielding around 5000 samples over 24 different resource configurations. In the following experiment, we use MNIST to train MobilenetV2\cite{sandler2018mobilenetv2}. 

\subsection{Numerical Evaluation of the Adaptive Latency Predictor}
To construct the dataset $D$ used for pre-training the latency predictor, we gather data from diverse devices, including CPUs such as Intel(R) Xeon(R) CPU E5-2640, AMD EPYC 7402, and Huawei Kunpeng 920. To improve the generalization performance of the predictor, we simulate various levels of computing capacities by the \textit{taskset} tool in Linux, which controls the number of processor cores executing a given process on the device. We measure the latency of all possible sub-models on each device, yielding around 5000 samples over 24 different resource configurations.

\begin{figure*}[htbp]
\centering
	  \subfloat[]{
      \includegraphics[height=1.45in]{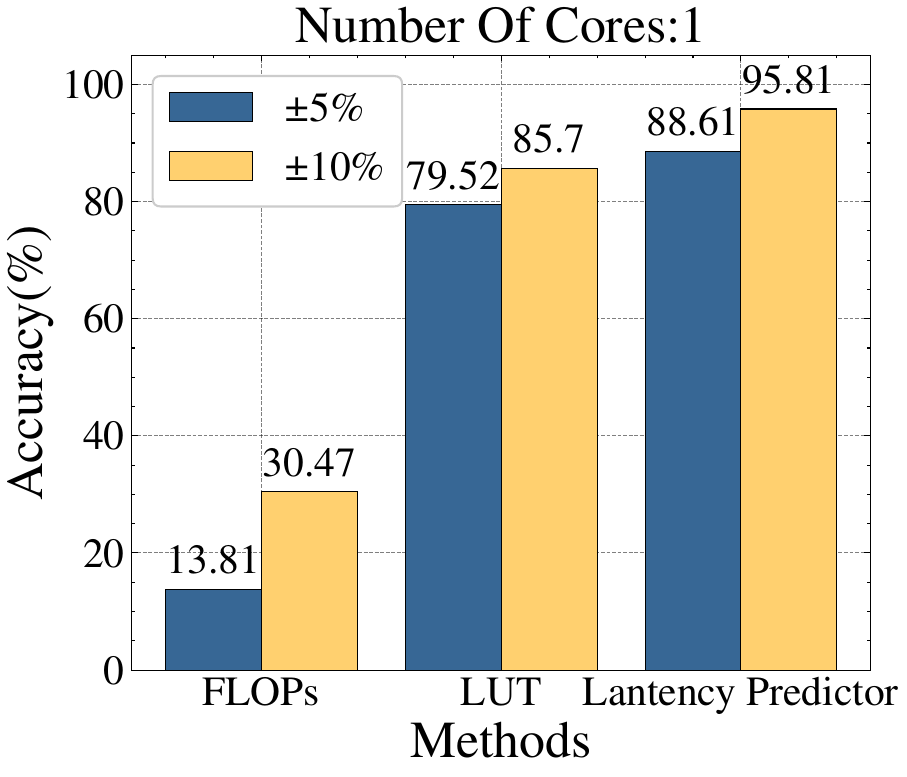}}
    \hfil
    \subfloat[]{
      \includegraphics[height=1.45in]{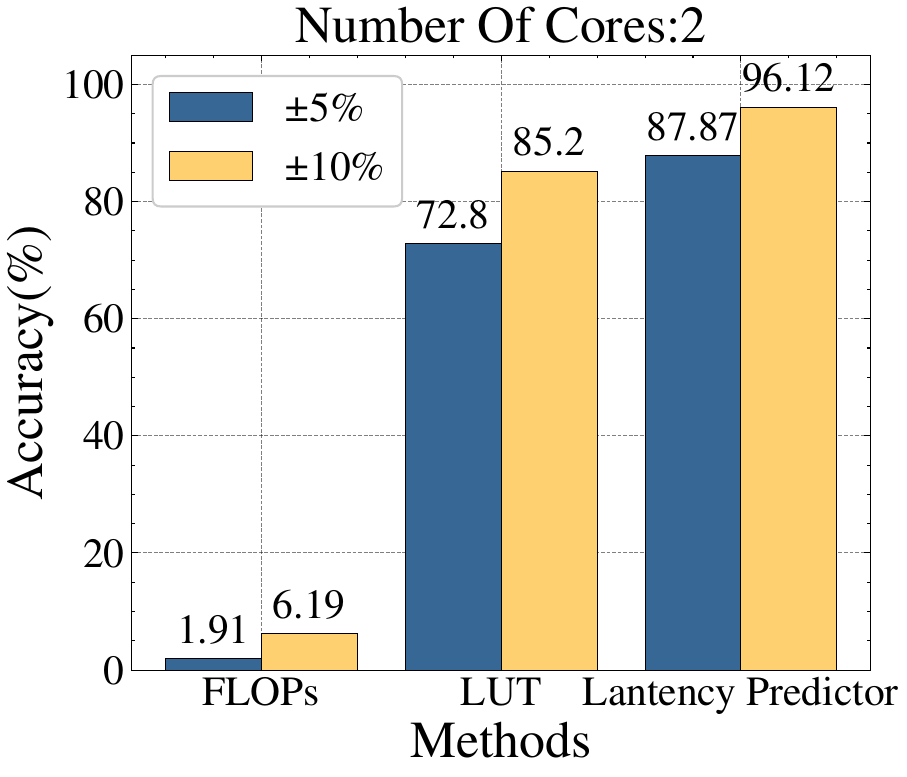}}
     \hfil
     \subfloat[]{
     \includegraphics[height=1.45in]{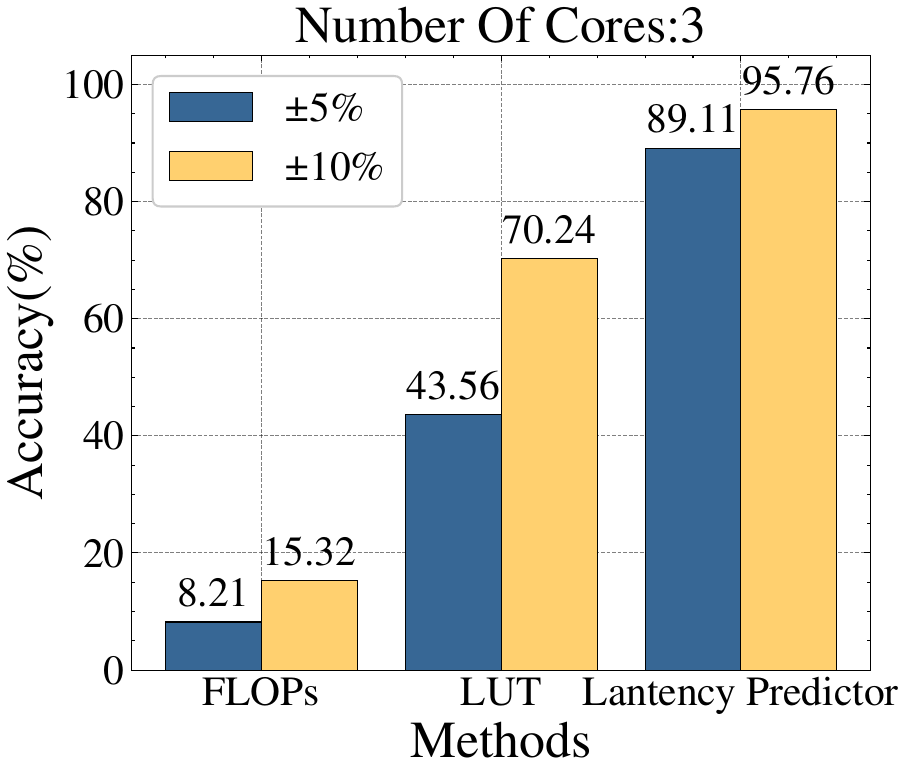}}
   \hfil
   \subfloat[]{
     \includegraphics[height=1.45in]{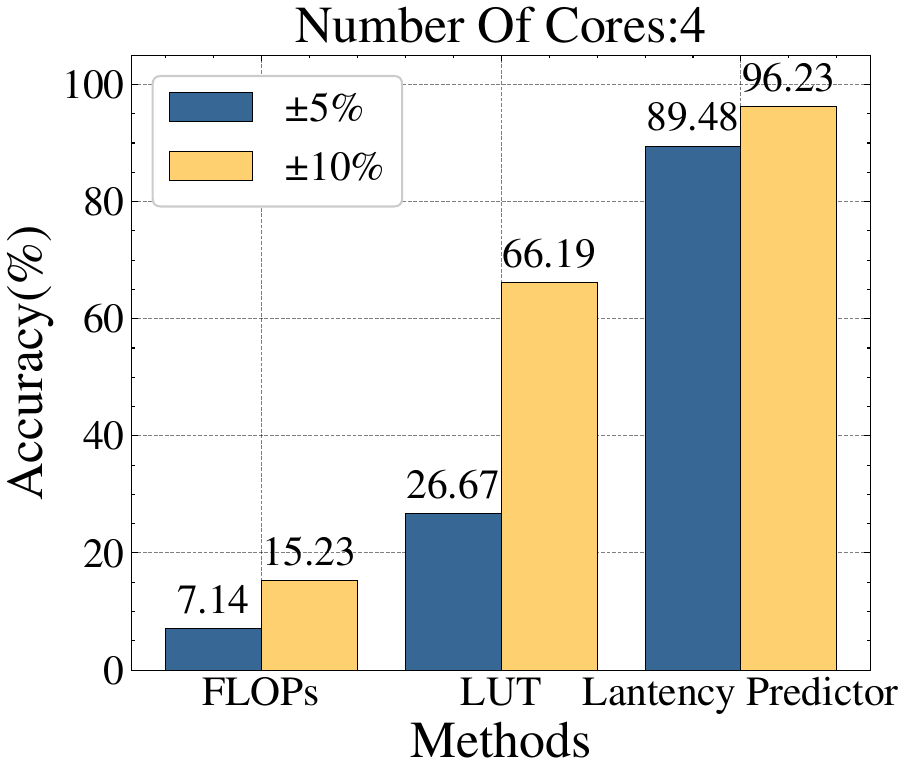}}
\captionsetup{justification=centering}
\caption{Performance comparisons between different latency prediction methods with different numbers of CPU cores used.}
\label{latency predictor performance}
\end{figure*}

\subsubsection{Prediction Accuracy Comparison}
We first evaluate the prediction accuracy of our latency predictor, where we use MNIST to train MobilenetV2\cite{sandler2018mobilenetv2} as the edge pipeline-parallel training task. We use the widely used FLOPs-based method\cite{he2018amc, li2016pruning} and loop up table(LUT)-based method\cite{wu2019fbnet, dai2019chamnet} as the baselines. The FLOPs-based method assumes that a model's FLOPs and its execution time are linearly correlated. Thus by utilizing this linear correlation, the execution latency of a model that has not been executed yet can be predicted. On the other hand, the LUT-based method involves pre-computing and storing the execution latency of each operator in a model into a table. It then looks up the execution latency of each operator from the table and sums up the overall execution latency of a model. For comparison, We take the Intel(R) Xeon(R) Gold 6133 as the unseen device that is not used in the training, and conduct experiments with its number of CPU cores limited to 1, 2, 3, and 4, respectively. We then employ FLOPs-based, LUT-based, and our own latency predictor method to predict the execution latency of every sub-model of MobileNetV2 in this evaluation. We also record actual execution latency to calculate the accuracy of the predictions. The accuracy is calculated based on the ±5\% and ±10\% error-bound of the latency, both of which are widely used in the literature\cite{dudziak2020brp}. Moreover, to ensure the correctness of the collected latency data, we measure the execution time of each sub-model 100 times and take the average. The experimental results are shown in Fig.~\ref{latency predictor performance}, with each sub-figure representing a different number of CPU cores. As it can be seen in Fig.~\ref{latency predictor performance}, the latency predicted by the FLOPs is significantly inaccurate under all resource conditions. The LUT-based prediction method shows a performance decline as the number of CPU cores increases, which can be attributed to that computation optimizations at the hardware or runtime level occur when more cores are used. Our latency predictor consistently outperforms the other two methods and demonstrates greater stability across all cases. It achieves an accuracy of approximately 88\% with a ±5\% error-bound and over 95\% with a ±10\% error-bound on average.

\begin{figure}[htbp]
\centerline{\includegraphics[width=3.5in]{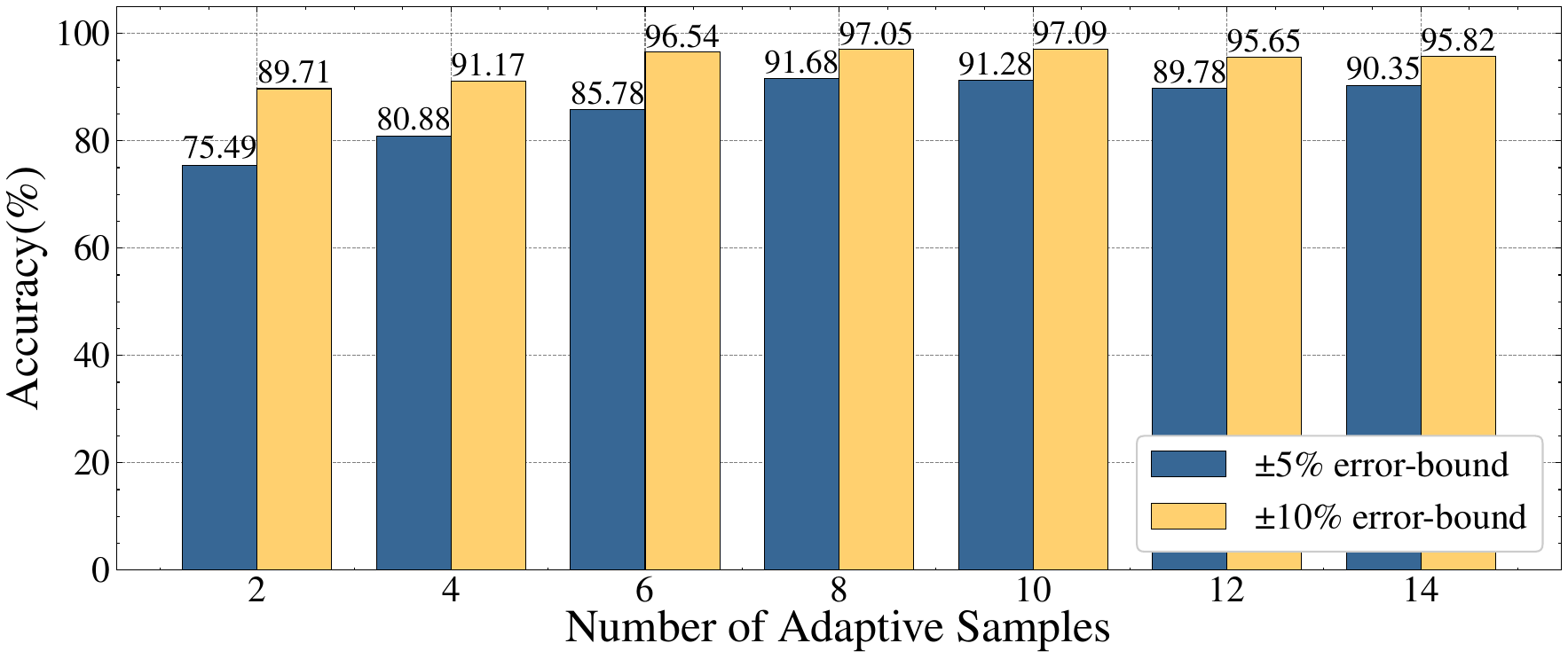}}
\caption{Evaluations of the prediction performance by the adaptively updated latency predictor with respect to different numbers of adaptive samples.}
\label{adaptive update performance}
\end{figure}

%\subsubsection{Evaluation of A}
Next, we evaluate the adaptive ability of the predictor, where we measure the prediction accuracy of the predictor after updating it by a new dataset $\hat{D}$. We adopt a cross-validation approach to avoid impact from different devices. For each iteration, we select one device as the unseen device and construct a new dataset $\hat{D}$ with it, while the other devices are used to generate the training dataset that pre-trains the latency predictor. We vary the total number of samples in the $\hat{D}$ from 2 to 14. These samples are selected from the hardware performance samples mentioned in \ref{subsec: predictor}. If there are no sufficient samples, we randomly sample the execution time of all sub-models and add the results to $\hat{D}$. The execution time of the sub-models that are not in $\hat{D}$ are then used as the test dataset. Then we combine the $\hat{D}$ with the pre-training dataset $D$ to continuously train the pre-trained latency predictor, and measure the prediction performance with the test dataset. We finally aggregate the results from all the iterations and calculate the average prediction accuracy, which is demonstrated in Fig.~\ref{adaptive update performance}.
% To evaluate the adaption ability of the predictor, we vary the number of samples in the adaptation dataset from 2 to 14 and measure the accuracy after incorporating the adaption samples into the training dataset. If there are insufficient samples, we randomly sample data without repetition from all sub-models and add it to the adaptation dataset. The remaining sub-models are then used for latency collection to verify the performance of the predictor. To prevent coincidences from specific device combinations, we adopt a cross-validation approach. For each iteration, we select one device as the unseen device to collect adaptation and test data. We then collect and use data from the remaining devices to train the latency predictor and verify its performance. Finally, we combine all results and calculated the average accuracy.

From the results shown in Fig.~\ref{adaptive update performance}, we can see that the performance of the latency predictor continuously improves as the number of samples in the new dataset $\hat{D}$ increases. The predictor eventually stabilizes around 90\% accuracy under a ±5\% error-bound and around 96\% accuracy under a ±10\% error-bound. This demonstrates that our latency predictor can be adapted to unseen devices with relatively few new training samples. Specifically, the results indicate that the performance of the latency predictor convergences after retraining the predictor with only eight new samples. However, in a relatively long period, the resources of each device, such as the device's battery status and the computation workloads, keep varying, which can affect the execution time of each sub-model. Hence, we periodically collect the sub-model's execution time from each device and use these newly collected data to update the latency predictor, which helps the predictor adapt to the time-varying resources on the devices.

\subsubsection{Acceleration by the Latency Predictor}
Next, we compare the edge pipeline parallel training time leveraging our latency predictor with two benchmarks, i.e., FTPipeHD and the edge pipeline parallel training with average partitioning. To enhance the reliability of the results, we consider three different scenarios for experiments where different devices are used as the central nodes and the work nodes, which are shown in Table~\ref{latency scenarios}.

\begin{table}[htbp]
\caption{Three scenarios with different device as the central node}
\scriptsize
\centering 
\begin{tabular}{cll}
    % \toprule
    \hline
    Scenario No. & Central Node(Num of cores) & Worker Node(Num of cores)\\
    % \midrule %[2pt]  
    \hline
    1 & Xeon E5-2640(2) & \tabincell{l}{Xeon Gold 6133(1) \\ Xeon Gold 6133(1)} \\
    \hline
    2 & Xeon E5-2640(3)  & \tabincell{l}{Kunpeng 920(1) \\ Xeon Silver 4210R(1)} \\
    \hline
    3 & AMD EPYC 7402(4) & \tabincell{l}{Xeon E5-2640(3) \\ Kunpeng 920(2) \\ Xeon Gold 6133(1)} \\
    % \bottomrule %[2pt]
    \hline
    &
\end{tabular}
\label{latency scenarios}
\end{table}

\begin{table}
\caption{The model partition points for different methods}
\scriptsize
\centering 
\begin{tabular}{lccc}
    \hline
    Method &\multicolumn{3}{c}{Partition Point}\\
    &Scenario 1&Scenario 2&Scenario 3 \\
    \hline
    Average Partition & (7,14)& (7,14)&(5,10,15) \\
    FTPipeHD & (8,14)& (10,13)&(8,12,16)\\
    Our Method& (7,13)& (11,15)&(8,13,16)\\
    \hline
    &
\end{tabular}
\label{latency partiton point}
\end{table}

\begin{table}
\scriptsize
\centering 
\begin{tabular}{lccc}
    \hline
    Method &\multicolumn{3}{c}{Training Time(min)}\\
    & Scenario 1&Scenario 2&Scenario 3\\
    \hline
    Average Partition &  44.44& 41.72& 27.07\\
    FTPipeHD & 42.26& 38.46& 26.22\\
    Our Method& 38.69& 35.42& 26.19\\
    \hline
    &
\end{tabular}
\label{latency training time}
\end{table}

We list the model partition points calculated by the three methods in Table.~\ref{latency partiton point} and show the training time for the first ten epochs in Table.~\ref{latency training time}. As can be observed from Table.~\ref{latency partiton point} and Table.~\ref{latency training time}, our latency predictor achieves a significant acceleration on training in Scenario 1 and Scenario 2, and a slight improvement in Scenario 3. This is because the overall computing capacities of the devices in Scenario 3 are much more than the other two scenarios, and there is still a surplus of computing resources after each device has allocated resources for training their local sub-models. Therefore, each device is able to execute its local sub-model in a relatively short period regardless of partitioning points. However, in scenarios where the overall computing capacities of devices are low or there is a large difference in computing capacity between devices, such as Scenario 1 and Scenario 2, it is important to balance the computing load in order to accelerate the overall training.

\subsection{Numerical Evaluation of the Bit-level Fast Compression Scheme}
Next, we evaluate the performance of the proposed bit-level fast compression scheme in accelerating pipeline training. We choose two representative DL applications, i.e., the image classification task and the human activity recognition task. For the image classification task, we use the CIFAR10 dataset and train a MobileNetV2 model. For the human activity recognition task, we use the UCI-HAR dataset\cite{anguita2013public} collected from smartphone's inertial sensors and train a simple five-layer Convolutional Neural Network (CNN) model proposed in \cite{chen2020fedhealth}. 

\subsubsection{Communication Overhead Comparison}
We first compare the size of the transmitted data before and after applying our proposed compression scheme, where the corresponding compression ratio (CR) is also calculated. The data compression results are shown in Table~\ref{transmission data comparison}. It can be observed from Table~\ref{transmission data comparison} that even for the layer in MobileNetV2 that outputs the minimum amount of data, it still needs to transmit 80.47KB if no compression is applied. The experimental results show that the proposed compression scheme achieves nearly a compression ratio of $\frac{32}{k}$ with $k$-bit quantization (i.e., the \emph{$k$-bit comp.}) for most cases. The gap between this theoretical compression ratio and the actual one is caused by transmitting the additional data, e.g. the coefficients $\boldsymbol{\alpha}$, and information needed for the distributed parallel training including the weight version and the batch id. This gap becomes larger if the size of the features is smaller since the size of the additional data and the distributed parallel training information are getting not negligible. Note that with quantization, we can only achieve a compression ratio of at most 4 even if we quantize a 32-bit floating point into an integer with less than 8 bits, since the minimum number of bits to represent a value in mainstream programming languages is 8. Moreover, if we use the MBQ without encoding, as shown in the \emph{$k$-bit w/o enc.}, the compression ratio is usually lower than 4. This is because each value is decomposed into $k$ binary bases and each binary value is stored and transmitted as an 8-bit integer. These results validate the significance of the proposed encoder in reducing the amount of transmitted data, i.e., communication overhead, together with the proposed quantizer.

\begin{table}[htbp]
\caption{The transmission data size and the compression ratio comparison}
\centering
\begin{threeparttable}
\resizebox{\linewidth}{!}{

\begin{tabular}{c|cc|cc|cc|cc}
\hline
 & \multicolumn{4}{c|}{Largest Latent Output} & \multicolumn{4}{c}{Smallest Latent Output} \\
 \hline
  & \multicolumn{2}{c|}{MobileNetV2}  & \multicolumn{2}{c|}{Simple CNN} & \multicolumn{2}{c|}{MobileNetV2} & \multicolumn{2}{c}{Simple CNN}\\
 \hline
 & \tabincell{c}{Size \\ (KB)} & CR & \tabincell{c}{Size \\ (KB)} & CR & \tabincell{c}{Size \\ (KB)} & CR & \tabincell{c}{Size \\ (KB)} & CR\\
 \hline
 Original & 4096 & 1.00 & 2016 & 1.00 & 80.47 & 1.00 & 25.45 & 1.00 \\
 \hline
 2-bit comp.\tnote{1} & \textbf{257.1} & \textbf{15.93} & \textbf{127.1} & \textbf{15.86} & \textbf{6.09} & \textbf{13.20} & \textbf{2.65} & \textbf{9.62}\\
 %\hline
 2-bit w/o enc.\tnote{2} & 2049 & 2.00 & 1009 & 2.00 & 41.08 & 1.96 & 13.57 & 1.88\\
 \hline
 3-bit comp. & 385.1 & 10.63 & 190.1 & 10.61 & 8.60 & 9.35 & 3.43 & 7.41\\
 %\hline
 3-bit w/o enc. & 3073 & 1.33 & 1513 & 1.33 & 61.08 & 1.32 & 19.82 & 1.28\\
 \hline
 4-bit comp. & 513.1 & 7.98 & 253.1 & 7.97 & 11.1 & 7.25 & 4.22 & 6.03\\
 %\hline 
 4-bit w/o enc. & 4097 & 1.00 & 2017 & 1.00 & 81.09 & 0.99 & 26.08 & 0.98\\
 \hline
 uniform comp.\tnote{3} & 512.5 & 7.99 & 252.5 & 7.98 & 10.51 & 7.66 & 3.62 & 7.02\\
 %\hline
 uniform w/o enc. & 1024 & 4.00 & 504.5 & 4.00 & 20.49 & 3.93 & 6.73 & 3.78\\
 \hline
 \end{tabular}
}
	\begin{tablenotes}
		\item[1] The $k$-bit comp. refers to the $k$-bit forward compression.
		\item[2] The $k$-bit w/o enc. refers to the $k$-bit forward quantizer without the \\ encoder.
		\item[3] The uniform comp. refers to the 4-bit uniform backward compression.
	\end{tablenotes}	
\end{threeparttable}
\label{transmission data comparison}
%\end{center}
\end{table}

\begin{figure}[htbp]
\centering
\subfloat[\label{latency_max}]{\includegraphics[height=1.20in]{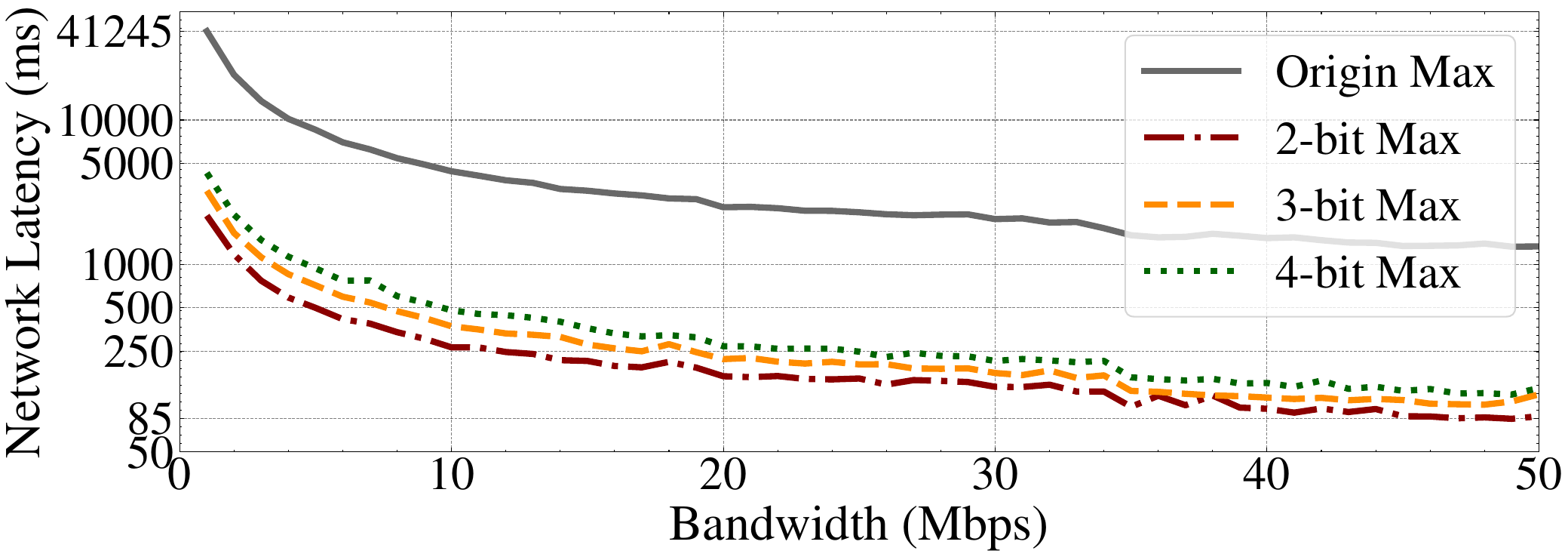}
}
\hfil
\subfloat[\label{latency_min}]{\includegraphics[height=1.20in]{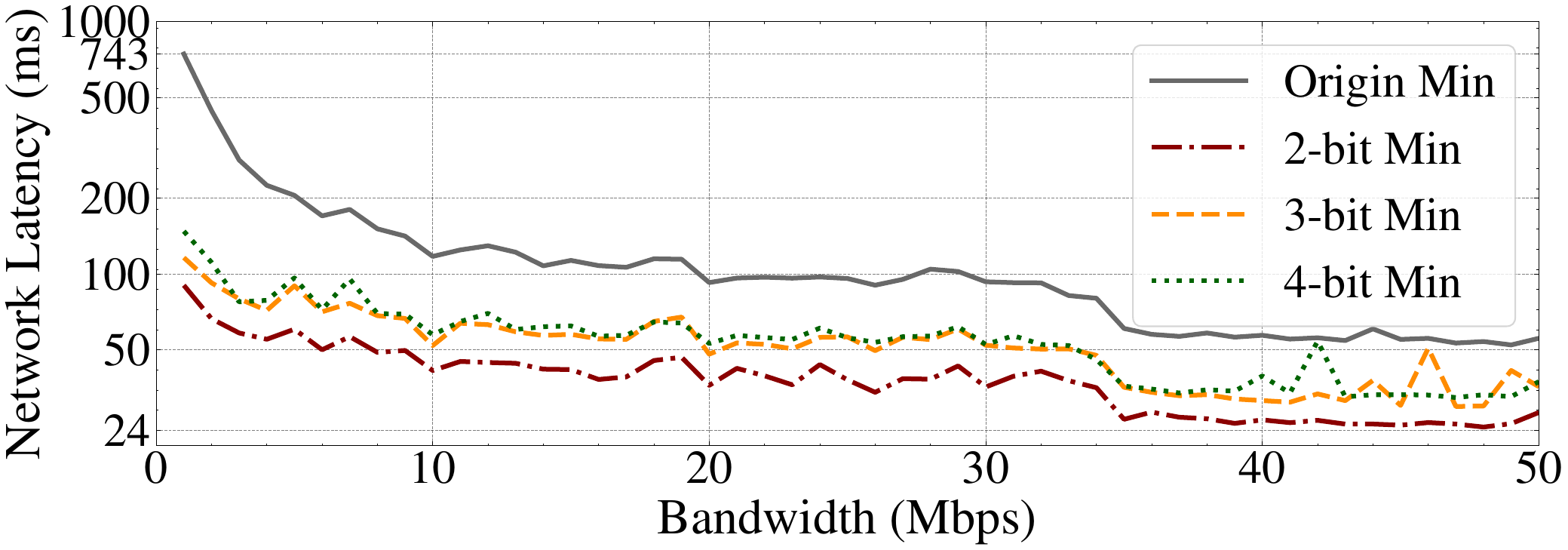}}
\caption{Comparison of the transmission network latency between the $k$-bit compression scheme and the no compression scheme on the MobileNetV2.(a) Network latency for the largest latent features. (b) Network latency for the smallest latent features.} 
\label{network latency}
\end{figure}

Next, we measure the actual network latency of transmitting the latent features of MobileNetV2. We conduct experiments by transmitting the latent features from a MacBook Pro to a desktop PC using a real WiFi connection. The network latency is measured as the time interval between the start of data transmission from the transmitting device and the successful reception of data by the receiving device. The WiFi network used is under a commonly used network bandwidth ranging from 1Mbps to 50Mbps, which is controlled by using the Network Link Conditioner tool in macOS. We conduct 50 measurements at each bandwidth and calculate the average network latency. Then we compare the latency between the data without compression and the data with the fixed $k$-bit compression, which is shown in Fig.~\ref{network latency} where \emph{Max} and \emph{Min} means sending the intermediate results with the maximal and minimum amount of latent features, respectively. As the figure shows, it still takes around 1300ms to transmit the largest latent features without compression even when the bandwidth is 50Mbps, while the proposed compression significantly reduces the latency to around 100ms. For the smallest latent features, when the bandwidth is 50Mbps, the transmission latency reduces from 55ms to about 25ms when 2-bit compression is used. We can also observe a significant reduction in latency across different bandwidths. We can note that the proposed compression significantly reduces the transmission network latency up to 1/13. 
We also note that the latencies by compression with different bit widths are getting closer with the increase of the bandwidth. Hence, when the available bandwidth is large, we can use 4-bit compression to improve the model accuracy. On the other hand, when the bandwidth is limited, we can use 2-bit compression to reduce the transmission latency.

%Finally, we note that the figure only shows the latency reduction of transmitting the feature of one batch. The total latency reduction during the training will be much greater than this. We validate this by deploying our system on the Raspberry Pi in Section~\ref{raspberry pi experiment}.} 

\subsubsection{Model Performance Evaluation}
\label{model performance evaluation}
Then, we evaluate the model performance under compression schemes with different bit widths and compare them with the scheme without compression in terms of model performance. We first compare our proposed MBQ scheme with other quantization approaches. Then we compare the performance of the proposed adaptive scheme with the fixed bit-width schemes. Note here we use three independent processes on the CPU to simulate three edge devices in the distributed parallel manner.

We first compare the proposed MBQ scheme with two quantization methods and the case with no quantization used. One baseline quantization method is the uniform quantization with the scaling factor $s$ initially set to 0.25 and updated by its gradients with respect to the training loss. The other is AMSGrad\cite{qu2020adaptive}, which prunes the number of $\alpha$ by computing the increment of the training loss incurred by each $\alpha$. For the proposed scheme and both baselines, the forward bit width $k$ is set to either 2, 3 or 4, while the backward bit width is set to 8 bits. The model accuracy on the test dataset is listed in Table~\ref{model performance with baseline}, where \emph{No compression} refers to FTPipeHD\cite{chen2023ftpipehd} without any data compression.

\begin{table}[htbp]
\caption{The model accuracy comparison under different quantization methods}
\centering
\begin{threeparttable}
\resizebox{\linewidth}{!}{
\begin{tabular}{c|cc|cc|cc}
\hline
 & \multicolumn{2}{c|}{2 Bits} & \multicolumn{2}{c|}{3 Bits} & \multicolumn{2}{c}{4 Bits}\\
 \hline
  & CIFAR10 & HAR & CIFAR10 & HAR & CIFAR10 & HAR \\
 \hline
No compression & 85.72\% & 97.47\% & 85.72\% & 97.47\% & 85.72\% & 97.47\% \\
\hline
\textbf{Proposed MBQ} & \textbf{85.23\%} & \textbf{94.93\%} & \textbf{85.09\%} & \textbf{96.04\%} & \textbf{85.57\%}  & \textbf{96.79\%} \\
\hline
Uniform Quant. & 84.66\% & 80.85\% & 84.38\% & 78.67\% & 83.77\% & 94.30\% \\
\hline
AMSGrad MBQ & 81.67\% & 92.84\% & 82.01\% & 87.02\% & 77.93\% & 93.90\%\\
\hline
 \end{tabular}

}
\end{threeparttable}
\label{model performance with baseline}
%\end{center}
\end{table}

As we can observe from Table~\ref{model performance with baseline}, the proposed MBQ has better testing accuracy than other quantization methods in both CIFAR10 and HAR. We have also found that the AMSGrad MBQ sometimes diverges the model during training, resulting in the degradation of model accuracy. When compared with FTPipeHD, the proposed MBQ method has a comparable testing accuracy on CIFAR10, while it suffers a slight accuracy drop on HAR due to the compression.  It also can be observed that as the bit width decreases, the testing accuracy drops on HAR. However, the testing accuracies under different bit widths on CIFAR10 remain approximately the same. This implies that the quantization may be more likely to degrade the model when the model is small.

Next, we evaluate the performance of the adaptive compression scheme, which is compared to the fixed bit-width and the scheme without compression. We consider three fixed bit-width compression cases with the bit width in the forward compression to be 2, 3 and 4, respectively, while  the bit width in backward compression is 8 in all three cases. For the adaptive scheme, we increase the forward bit width at epoch 130 and 175 in the MobileNetV2 and epoch 60 and 160 in the simple CNN model from 2 to 4, respectively. We train both the MobileNetV2 and the simple CNN for a total 200 epochs. Hence we have the average forward bit width for the training of the two models to be 2.48 and 2.9, respectively. We also decrease the backward bit width at epoch 160 for the MobileNetV2 model and at epoch 120 for the simple CNN model from 8 to 4.  Similarly, the average backward bit width is 7.2 and 6.4. 
%We measure the model accuracy of the testing dataset on both CIFAR10 and HAR.

\begin{figure}[htbp]
\centering
     \includegraphics[width=3in]{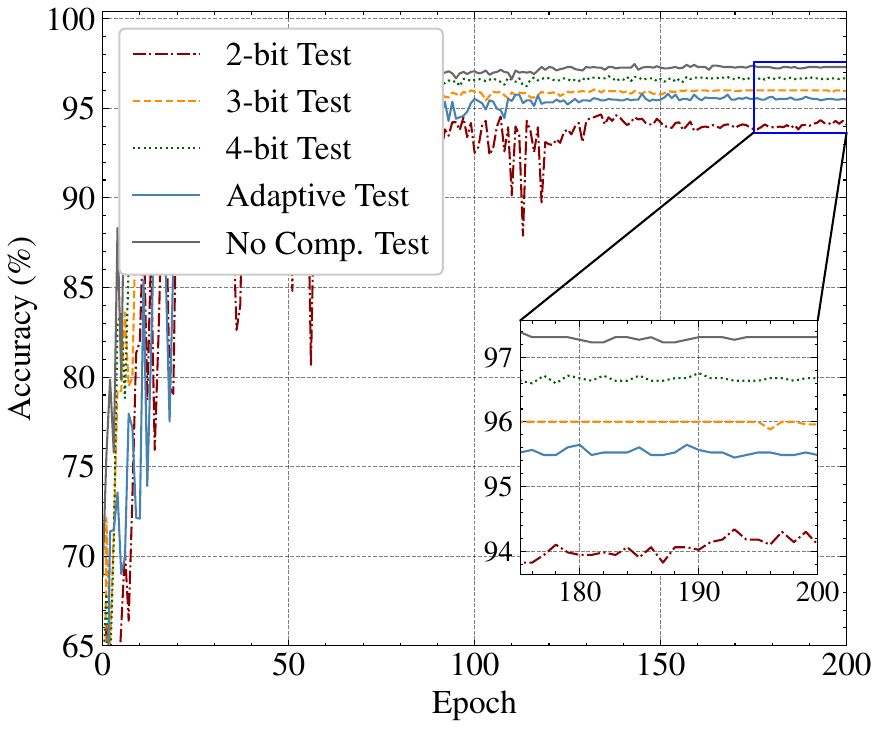}
\caption{Comparison between the adaptive compression scheme, fixed $k$-bit scheme and the no compression scheme in terms of model accuracy on the HAR dataset.}
\label{har_adaptive}
\end{figure}

\begin{figure}[htbp]
\centering
     \includegraphics[width=3in]{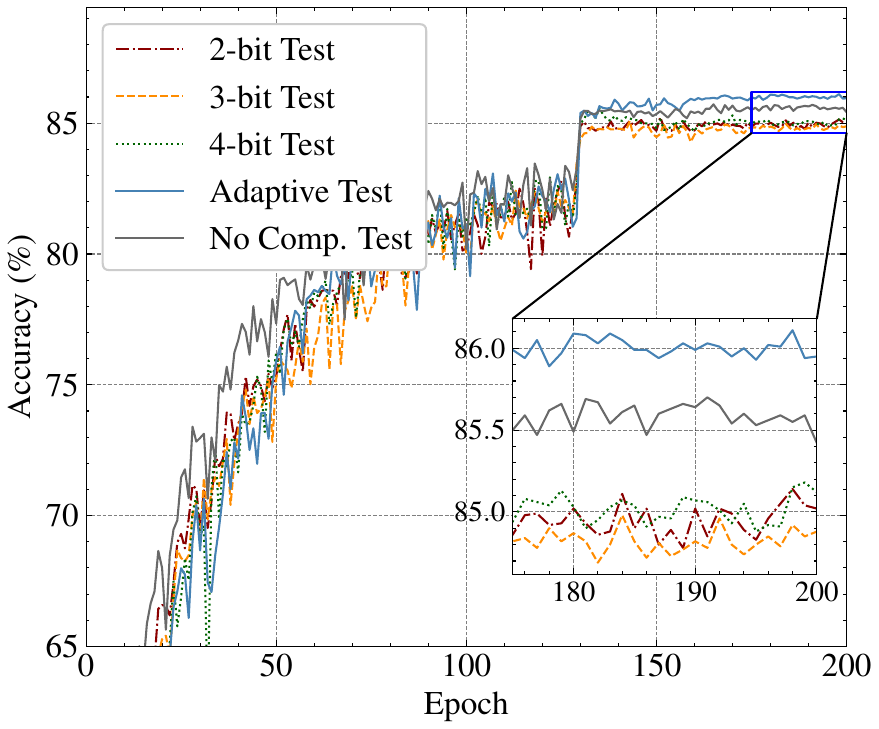}
\caption{Comparison between the adaptive compression scheme, fixed $k$-bit scheme and the no compression scheme in terms of model accuracy on CIFAR10 dataset.}
\label{mobilenet_adaptive}
\end{figure}

Fig.~\ref{har_adaptive} and Fig.~\ref{mobilenet_adaptive} illustrate the model accuracy on HAR and CIFAR10 respectively. It can be observed that all the accuracy curves fluctuate greatly during the training due to the parallel training manner and the quantization, however, becomes stable as the model converges. For the HAR dataset, all the schemes with compression suffer from performance degradation in terms of accuracy compared to the scheme without compression. As for the CIFAR10 dataset, however, the adaptive compression scheme  even outperforms the one without compression. This may be due to the fact that 2-bit compression introduces more noise to the data at the beginning of the training, which can be beneficial for the training. Moreover, as the learning rate decreases, a higher bit width can help the convergence of the model.

\subsection{Numerical Evaluation of the AccEPT Scheme}
Finally, we evaluate the performance of the AccEPT scheme on accelerating the edge pipeline parallel training, and compare to the cases with only latency predictor (LP) applied or only bit-level compression (BLC) applied, and FTPipeHD. We use Scenario 1 and 2 from Table~\ref{latency scenarios} as the experimental settings and train MobileNetV2 on the MNIST dataset as the training task, with the epochs set to 100. We also measure the training time under bandwidths ranging from 0.4Mbps to 4Mbps between devices. The evaluation results are demonstrated in Table~\ref{accept evaluation}.

From the Table.~\ref{accept evaluation}, we can see that the AccEPT scheme outperforms other methods in terms of the training time, with AccEPT achieving acceleration ratios of at most 3.01 compared to FTPipeHD. Under low-bandwidth conditions where the communication latency between devices dominates the total training time, our proposed compression greatly reduces the amount of information to be transmitted and significantly accelerates the training, with BLC achieving acceleration ratios of 2.84 and 2.07 compared to FTPipeHD in Scenarios 1 and 2, respectively. As the bandwidth increases, the communication latency decreases, and computation time becomes the main part in slowing down training. At this point, the effect of the latency predictor becomes obvious, with LP achieving acceleration ratios of 1.15 and 1.10 in Scenarios 1 and 2 at 4Mbps compared to FTPipeHD, respectively. 

\begin{table}[htbp]
\centering 
\caption{Comparison between AccEPT and other Methods on Training Time (Min)}
\begin{threeparttable}
\resizebox{\linewidth}{!}{
\begin{tabular}{ccccccc}
\hline
\multirow{2}{*}{Scenario\#} & \multicolumn{1}{c}{\multirow{2}{*}{Method}} & \multicolumn{5}{c}{Bandwidth(Mbps)}           \\
                             & \multicolumn{1}{c}{}   & 0.25    & 0.5     & 1       & 2      & 4      \\
\hline
\multirow{5}{*}{1}           & FTPipeHD                & 3439.17 & 1755.93 & 894.41  & 620.03 & 488.85 \\
&            BLC\tnote{1} &  1212.64 &952.16      &   546.26&    577.76&      466.341\\
&            LP\tnote{2} &    2326.17&     1280.96&     628.88&      455.57&423.07\\
& \textbf{AccEPT}             & \textbf{1139.95} & \textbf{711.25}  & \textbf{437.19}  & \textbf{423.59} & \textbf{403.75} \\
\hline
\multirow{5}{*}{2}           & FTPipeHD                & 2295.11 & 1235.48 & 621.758 & 484.66 & 397.4  \\
&            BLC &   1110.78&     721.68&    451.04&    442.07&      390.12\\
&            LP &   2069.48&     1178.71&    584.86&    400.83&      360.69\\
        & \textbf{AccEPT}  & \textbf{967.86}  & \textbf{639.66}  & \textbf{420.15}  & \textbf{364.24} & \textbf{351.41} \\
\hline
    &
\end{tabular}
}
\begin{tablenotes}
    \item[1] Only the bit-level compression(BLC) is implemented.
    \item[2] Only the latency predictor(LP) is implemented.
\end{tablenotes}
\end{threeparttable}
\label{accept evaluation}
\end{table}

% \begin{table}[htbp]
% \centering 
% \caption{Training Time (Min) Comparison On Raspberry Pi}
% \begin{threeparttable}
% \resizebox{\linewidth}{!}{
% \begin{tabular}{cccccc}
% \hline
% \multicolumn{1}{c}{\multirow{2}{*}{Method}} & \multicolumn{5}{c}{Bandwidth(Mbps)}           \\    & 0.25    & 0.5     & 1       & 2      & 4      \\
% \hline
% FTPipeHD                & 304.25  & 204.38 & 137.14  & 128.12 & 100.99 \\
% BLC\tnote{1}            &  221.31 & 151.69  & 117.78  & 114.29 & 99.81\\
% LP\tnote{2}             &  284.94 & 199.34 & 132.47  & 111.64 & 93.19 \\
% \textbf{AccEPT}         & \textbf{199.22} & \textbf{148.89}  & \textbf{114.62}  & \textbf{101.81} & \textbf{92.11} \\
% \hline
%     &
% \end{tabular}
% }
% \begin{tablenotes}
%     \item[1] Only the bit-level compressor(BLC) is implemented.
%     \item[2] Only the latency predictor(LP) is implemented.
% \end{tablenotes}
% \end{threeparttable}
% \label{accept evaluation on raspberrypi}
% \end{table}

\begin{figure}[htbp]
\centering
	  \subfloat[\label{raspberry}]{
      \includegraphics[width=3.35in]{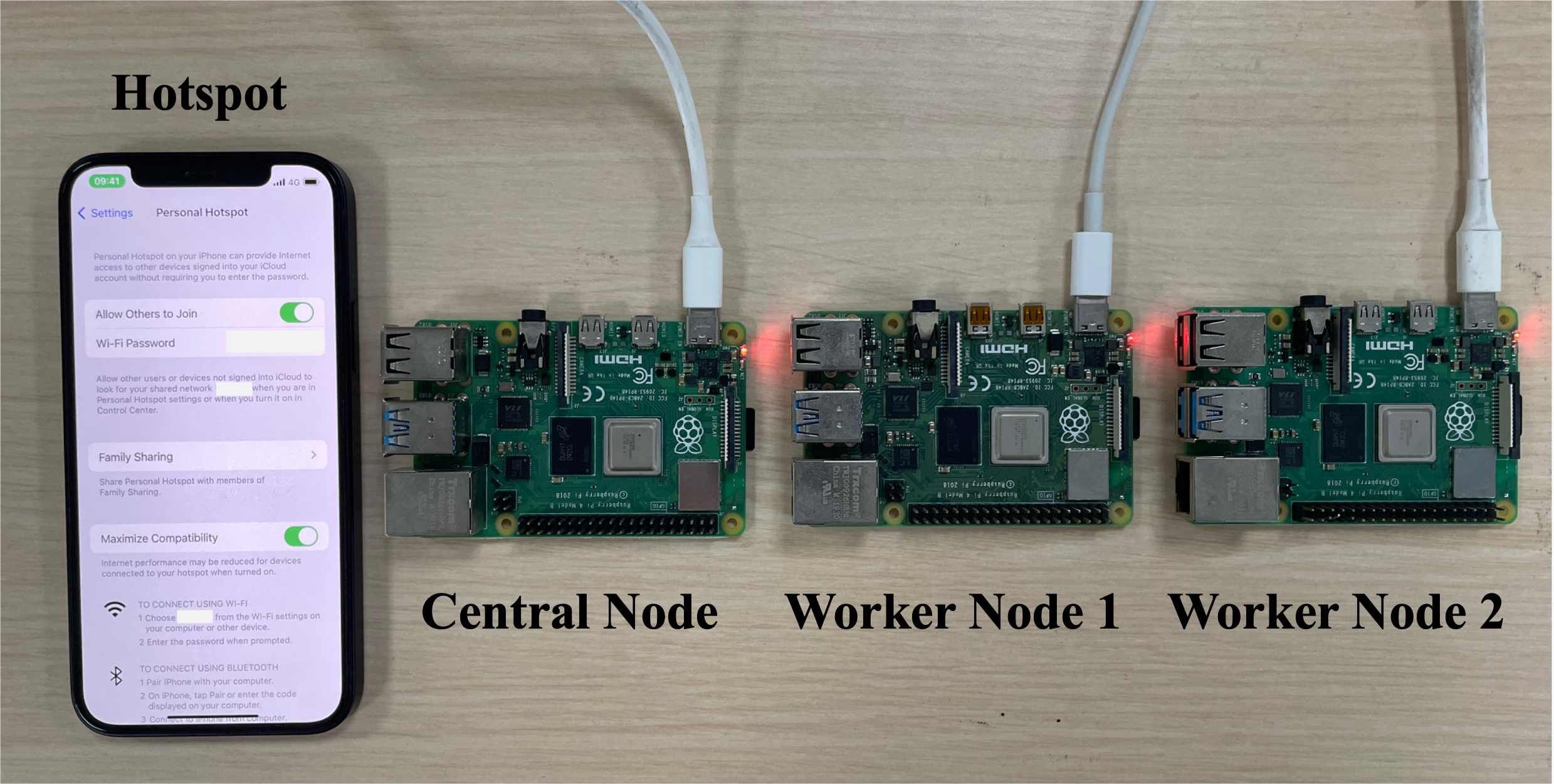}}
    \hfill
    \subfloat[\label{rasp_time}]{
      \includegraphics[width=3.48in]{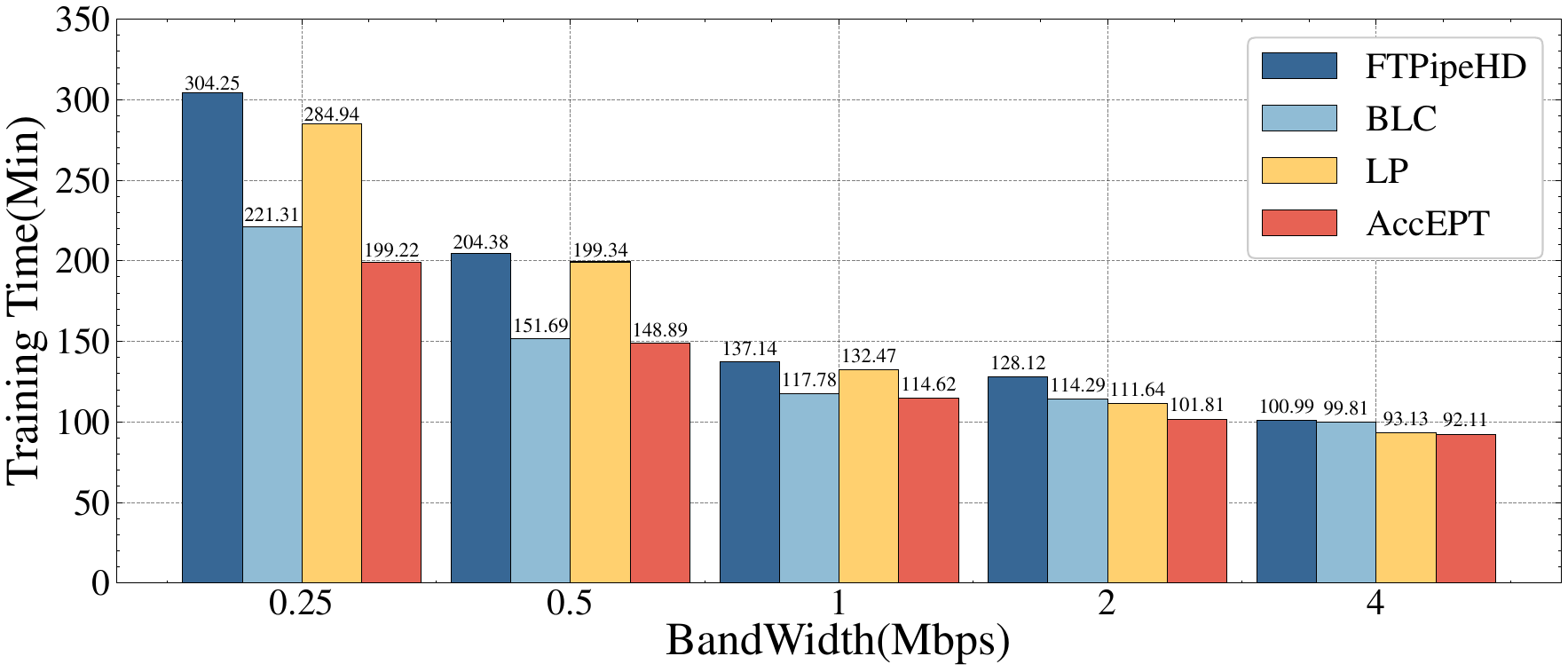}}
    \hfill
\caption{Training time evaluation on edge devices: (a) The Raspberry Pi used for evaluation; (b) Training Time Comparison between different schemes.}
\label{loss and time}
\end{figure}

We also deploy our proposed AccEPT to train the simple CNN model on the HAR dataset with three Raspberry Pi Model 4B, which are widely used edge devices. We note that due to the memory limitation of the Raspberry Pi, the training of the MobileNetV2 is not considered here. To simulate the heterogeneous devices scenario, we use \textit{taskset} tool to limit the number of CPU cores used in two devices to 2, where the rest one to 4. The three edge devices communicate with each other through a wireless local area network(LAN) provided by a mobile phone, where we conduct the experiments under different bandwidths ranging from 0.25Mbps to 4Mbps, which is controlled by the \textit{tc} tool. The actual deployment is shown in Fig.~\ref{raspberry}. Similarly, we compare the training time between the AccEPT, the cases with only the bit-level compression or only the latency predictor and the FTPipeHD, which is shown in Fig.~\ref{rasp_time}. 

It can be observed from Fig.~\ref{rasp_time} that although the size of transmission data is relatively small and the possible partition points are limited in the simple CNN compared with the one in the MobileNetV2, AccEPT still provides a significant training acceleration. Similar to the results in Table.~\ref{accept evaluation}, when the bandwidth is low, the time for transmitting the data between devices dominates the training time, making the data compression contribute more to the acceleration compared with the latency predictor. As the bandwidth increases, the training time of the case with only the latency predictor applied becomes smaller than that with only the bit-level compression, demonstrating that the latency predictor becomes more important when the transmission time decreases. We also note that due to the relatively small amount of transmitted data in the Simple CNN, the benefits of using the data compression are getting less significant with the increase of bandwidth. The evaluation results on both the edge servers and the Raspberry Pi devices validate the effectiveness of the proposed AccEPT scheme in accelerating pipeline training under different conditions.

\section{Conclusion}
\label{sec: conclusion}
In this paper, we proposed an acceleration scheme to speed up the distributed pipeline parallel training, called AccEPT. We first introduced an adaptive light-weight latency predictor to accurately predict the sub-model's execution time on each participating device, which helps to better partition the model across devices. Then we proposed a bit-level compression scheme to reduce the data size of the latent features transmitted during training, which decreases the communication overhead. In these two ways, the pipeline parallel training can be significantly accelerated compared to the existing methods, as demonstrated by the experimental results on both the edge servers and the Raspberry Pi devices.
%Based on the accurate latency prediction and reduced amount of transmitted data, the model can be better partitioned compared with the existing methods. We have conducted experiments to evaluate the performance of the latency predictor and the bit-level compressor separately from several aspects.  both the edge servers and the Raspberry Pi devices to show its acceleration performance.

\bibliographystyle{IEEEtran}
\bibliography{ref.bib}{}

\begin{IEEEbiography}[{\includegraphics[width=1in,height=1.25in,clip,keepaspectratio]{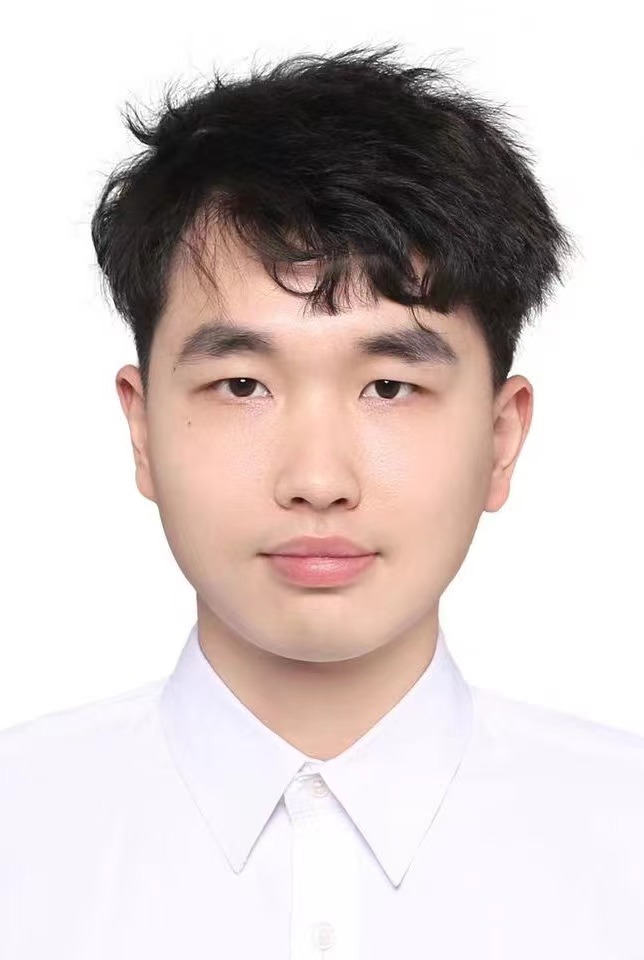}}]{Yuhao Chen}
received the Bachelor of Engineering degree from the College of Information Science and Electronic Engineering, Zhejiang University, in 2019. He is currently a Ph.D. candidate in cyberspace security at the College of Control Science and Engineering, Zhejiang University. His research interests include edge computing systems in IoT and semantic communication.
\end{IEEEbiography}

\begin{IEEEbiography}[{\includegraphics[width=1in,height=1.25in,clip,keepaspectratio]{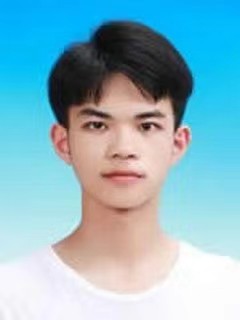}}]{Yuxuan Yan}
obtained his Bachelor of Engineering degree from the College of Information Science and Electronic Engineering at Zhejiang University in 2023. He is currently pursuing a Master's degree in the College of Information Science and Electronic Engineering at Zhejiang University, specializing in the field of edge computing and semantic communication.
\end{IEEEbiography}

\begin{IEEEbiography}[{\includegraphics[width=1in,height=1.25in,clip,keepaspectratio]{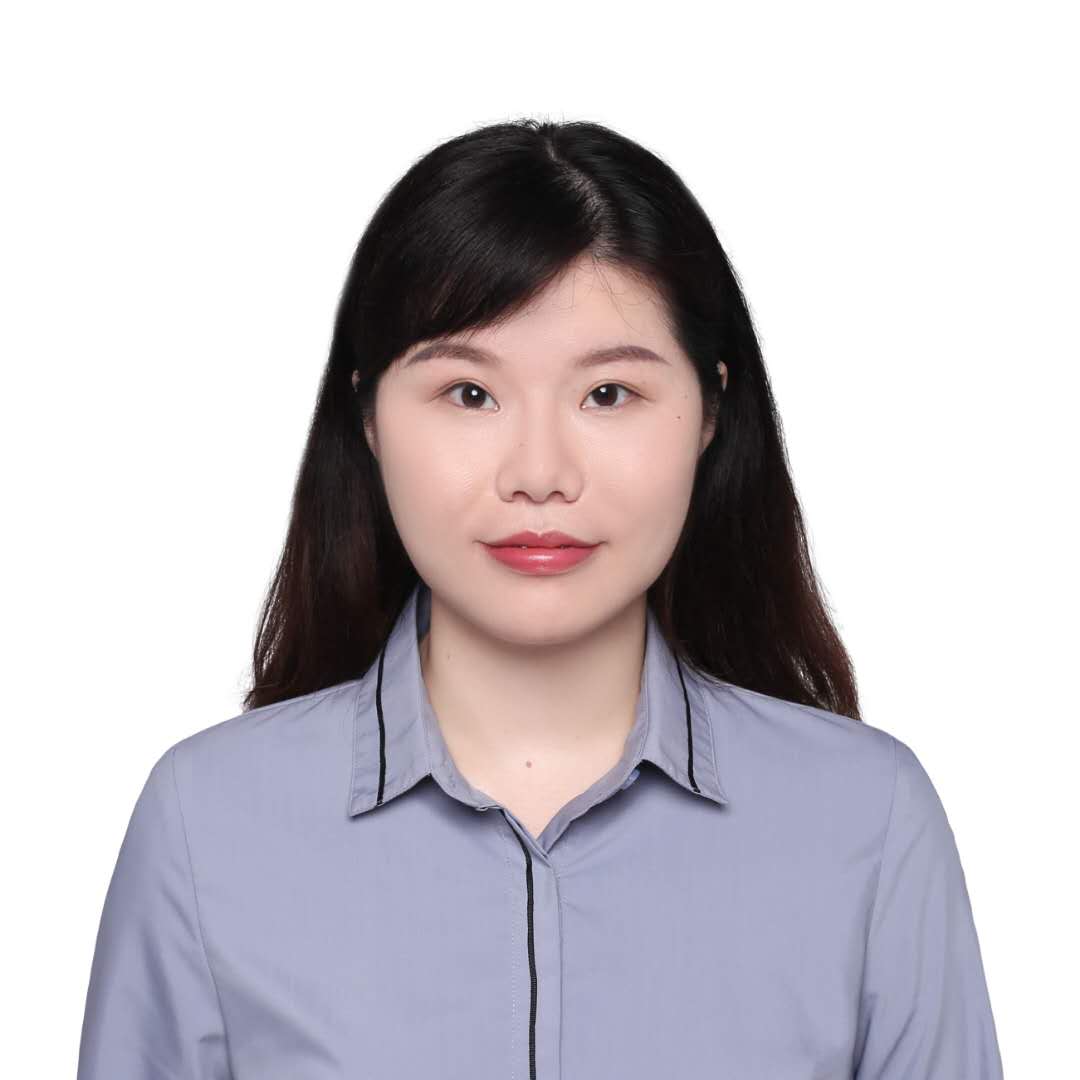}}]{Qianqian Yang}
(Member, IEEE) received the B.Sc. degree in automation from Chongqing University, Chongqing, China, in 2011, the M.S. degree in control engineering from Zhejiang University, Hangzhou, China, in 2014, and the Ph.D. degree in electrical and electronic engineering from Imperial College London, U.K. She has held visiting positions at CentraleSupelec in 2016 and the New York University Tandon School of Engineering from 2017 to 2018. After her Ph.D., she served as a Post-Doctoral Research Associate for Imperial College London, and as a Machine Learning Researcher for Sensyne Health Plc. 

She is currently a Tenure-Tracked Professor with the Department of Information Science and Electronic Engineering, Zhejiang University, China. Her main research interests include wireless communications, information theory,  machine learning and medical imaging. She serves as a Reviewer for IEEE TRANSACTIONS ON INFORMATION THEORY, IEEE TRANSACTIONS ON COMMUNICATIONS, IEEE TRANSACTIONS ON WIRELESS COMMUNICATIONS, etc.
\end{IEEEbiography}

\begin{IEEEbiography}[{\includegraphics[width=1in,height=1.25in,clip,keepaspectratio]{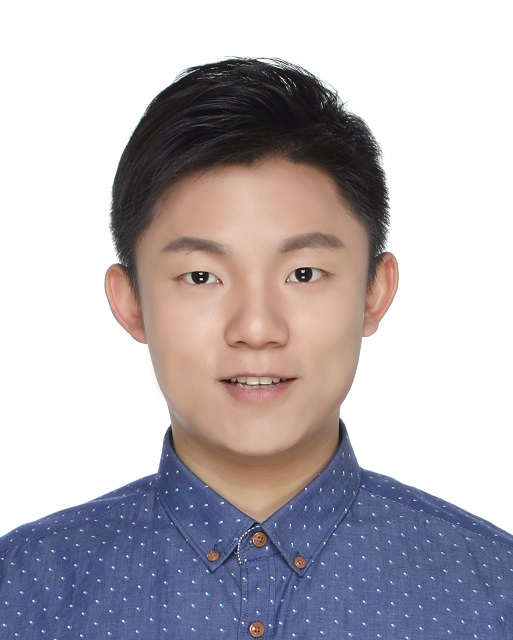}}]{Yuanchao Shu}
is currently a Qiushi Professor with the College of Control Science and Engineering at Zhejiang University, China. Prior to joining academia, he was a Principal Researcher with the Office of the CTO, Microsoft Azure for Operators, and the Mobility and Networking Research Group at Microsoft Research Redmond. His research interests lie broadly in mobile, sensing and networked systems. His previous research results have led to over 60 publications at top-tier peer-reviewed conferences and journals including ACM MobiCom, MobiSys, SenSys, UbiComp, USENIX Security, NSDI, JSAC, TMC, TWC, TPDS, TKDE etc. Dr. Shu currently serves on the editorial board of IEEE Transactions of Wireless Communications, ACM Transactions on Sensor Networks, and was a member of the organizing committee and TPC of conferences including MobiCom, MobiSys, SenSys, SEC, Globecom, ICC etc. He won five Best Paper/Demo (Runner-Up) Awards, and was the recipient of ACM China Doctoral Dissertation Award (2/yr) and IBM PhD Fellowship. Dr. Shu received his Ph.D. from Zhejiang University, and was also a joint Ph.D. student in the EECS Department at the University of Michigan, Ann Arbor. Dr. Shu is a senior member of ACM and IEEE.
\end{IEEEbiography}

\begin{IEEEbiography}[{\includegraphics[width=1in,height=1.25in,clip,keepaspectratio]{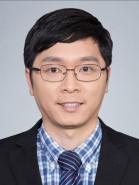}}]{Shibo He}
(M'13-SM'19) received the Ph.D. degree in control science and engineering from Zhejiang University, Hangzhou, China, in 2012. He is currently a Professor with Zhejiang University. He was an Associate Research Scientist from March 2014 to May 2014, and a Postdoctoral Scholar from May 2012 to February 2014, with Arizona State University, Tempe, AZ, USA. From November 2010 to November 2011, he was a Visiting Scholar with the University of Waterloo, Waterloo, ON, Canada. 

His research interests include Internet of Things, crowdsensing, big data analysis, etc. Prof. He serves on the editorial board for the IEEE TRANSACTIONS ON VEHICULAR TECHNOLOGY, Springer Peer-to-Peer Networking and Application and KSII Transactions on Internet and Information Systems, and is a Guest Editor for Elsevier Computer Communications and Hindawi International Journal of Distributed Sensor Networks. He was a Symposium Co-Chair for the IEEE GlobeCom 2020 and the IEEE ICC 2017, TPC Co-Chair for i-Span 2018, a Finance and Registration chair for ACM MobiHoc 2015, a TPC Co-Chair for the IEEE ScalCom 2014, a TPC Vice Co-Chair for ANT 2013´lC2014, a Track Co-Chair for the Pervasive Algorithms, Protocols, and Networks of EUSPN 2013, a Web Co-Chair for the IEEE MASS 2013, and a Publicity Co-Chair of IEEE WiSARN 2010, and FCN 2014.
\end{IEEEbiography}

\begin{IEEEbiography}[{\includegraphics[width=1in,height=1.25in,clip,keepaspectratio]{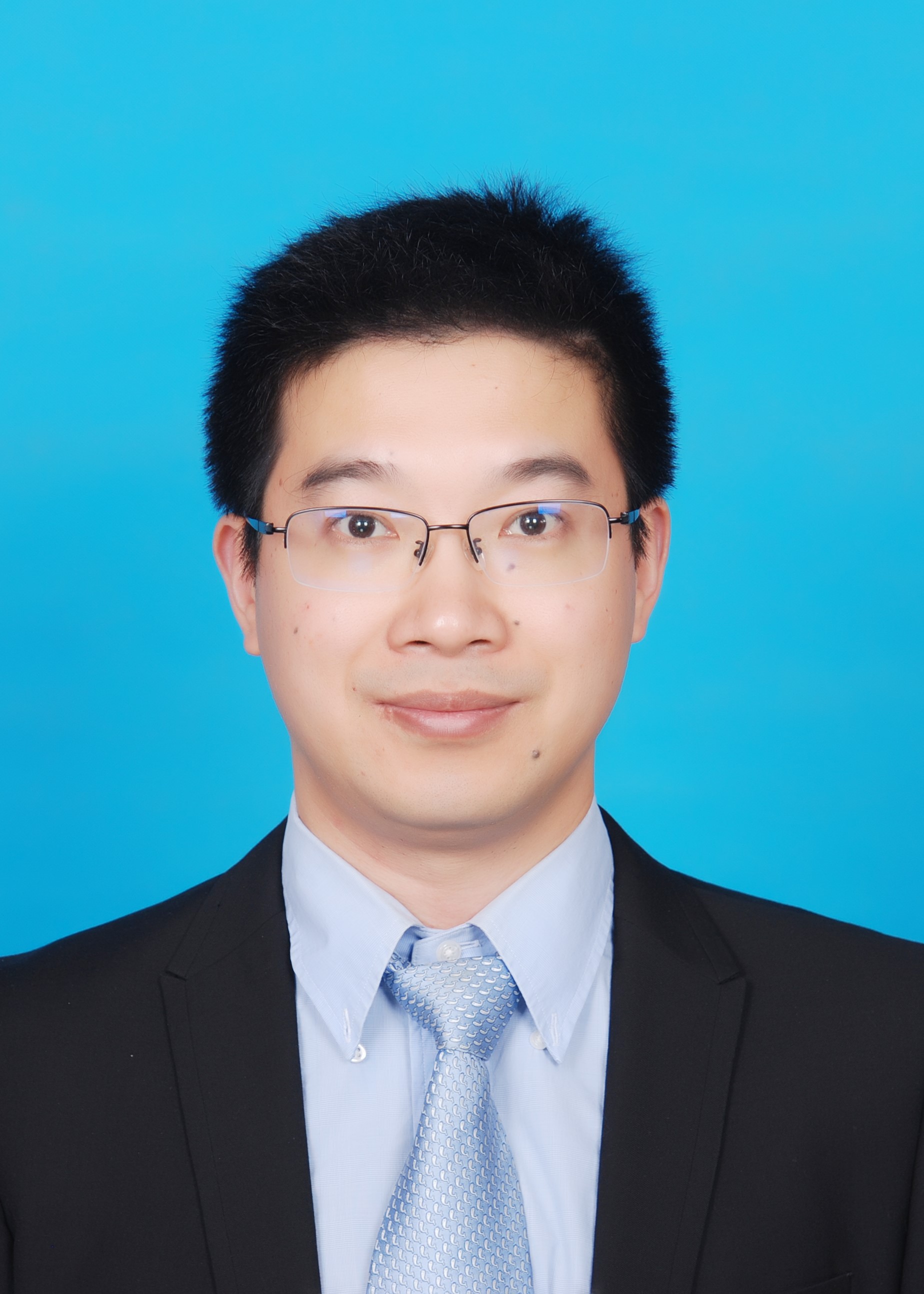}}]{Zhiguo Shi}
(M'10-SM'15) received the B.S. and Ph.D. degrees in electronic engineering from Zhejiang University, Hangzhou, China, in 2001 and 2006, respectively. Since 2006, he has been a Faculty Member with the Department of Information and Electronic Engineering, Zhejiang University, where he is currently a Full Professor. From 2011 to 2013, he visited the Broadband Communications Research Group, University of Waterloo, Waterloo, ON, Canada. His current research interests include signal and data processing. 

Prof. Shi serves as an Editor for the IEEE Network, IEEE Transactions on Vehicular Technology, IET Communications, Journal of The Franklin Institute. He was a recipient of the Best Paper Award from the ISAP 2020, IEEE GLOBECOM 2019, IEEE iThings 2019, EIA MILCOM 2018, IEEE WCNC 2017, CSPS 2017, IEEE/CIC ICCC 2013, IEEE WCNC 2013, IEEE WCSP 2012. He was also a recipient of first prize for Wu Wenjun AI Science and Technology Award in 2020, first prize for Science and technology Development of the Ministry of Education in 2015 and second prize for the Scientific and Technological Award of Zhejiang Province in 2012.

\end{IEEEbiography}

\begin{IEEEbiography}[{\includegraphics[width=1in,height=1.25in,clip,keepaspectratio]{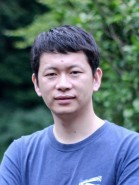}}]{Jiming Chen}
(Fellow, IEEE) received the PhD degree in control science and engineering from Zhejiang University, Hangzhou, China, in 2005. He is currently a professor with the Department of Control Science and Engineering, the vice dean of the Faculty of Information Technology, Zhejiang University. His research interests include IoT, networked control, wireless networks. He serves on the editorial boards of multiple IEEE Transactions, and the general co-chairs for IEEE RTCSA’19, IEEE Datacom’19 and IEEE PST’20. He was a recipient of the 7th IEEE ComSoc Asia/Pacific Outstanding Paper Award, the JSPS Invitation Fellowship , and the IEEE ComSoc AP Outstanding Young Researcher Award. He is an IEEE VTS distinguished lecturer. He is a fellow of the CAA.
\end{IEEEbiography}

\end{document}